\documentclass[final]{cvpr}

\usepackage{titling}
\usepackage{cvprtitle} 
\usepackage{times}
\usepackage{epsfig}
\usepackage{graphicx}
\usepackage{amsmath}
\usepackage{amssymb}
\usepackage{bm}
\usepackage{multibib}

\usepackage{bbding}
\usepackage{tabularx}
\usepackage{booktabs}
\usepackage{multirow}
\usepackage[keeplastbox]{flushend}
\usepackage{newtxtt}
\usepackage{etoolbox,siunitx}
\robustify\bfseries
\usepackage{color}
\definecolor{mybar}{rgb}{1.0, 0.4, 0.0}
\newcommand\cbar[3][mybar]{\colorbox{#1}{\color{black}\framebox(#2,#3){}}}

\definecolor{mygreen}{rgb}{0.2, 0.7, 0.1}
\definecolor{turquoise}{rgb}{0.173, 0.627, 0.537}

\usepackage[pagebackref=true,breaklinks=true,colorlinks,citecolor={mygreen},bookmarks=false]{hyperref}

\usepackage[capitalize]{cleveref}
\crefname{section}{Sec.}{Section}

\usepackage[format=plain,labelformat=simple,labelsep=period,font=small,skip=4pt,compatibility=false]{caption}
\usepackage[font=footnotesize,skip=2pt,subrefformat=parens]{subcaption}

\usepackage{pifont} 
\newcommand{\cmark}{\ding{51}}%

\usepackage[inline]{enumitem}
\usepackage{listings}

\definecolor{codegreen}{rgb}{0,0.6,0}
\definecolor{codegray}{rgb}{0.5,0.5,0.5}
\definecolor{codepurple}{rgb}{0.58,0,0.82}
\definecolor{backcolour}{rgb}{0.95,0.95,0.92}

\lstdefinestyle{mystyle}{
    backgroundcolor=\color{backcolour},
    commentstyle=\color{codegreen},
    keywordstyle=\color{magenta},
    numberstyle=\tiny\color{codegray},
    stringstyle=\color{codepurple},
    basicstyle=\ttfamily\footnotesize,
    breakatwhitespace=false,
    breaklines=true,
    captionpos=b,
    keepspaces=true,
    numbers=left,
    numbersep=5pt,
    showspaces=false,
    showstringspaces=false,
    showtabs=false,
    tabsize=2
}

\usepackage{xspace}

\renewcommand{\cf}{\emph{cf.}\@\xspace}
\newcommand{\Eq}{Eq.\@\xspace}

\newcommand*{\myparagraph}[1]{\smallskip\noindent\textbf{#1}\hspace{0.5em}}

\DeclareMathOperator*{\argmax}{arg\,max}

\newcommand\blfootnote[1]{%
  \begingroup
  \renewcommand\thefootnote{}\footnote{#1}%
  \addtocounter{footnote}{-1}%
  \endgroup
}

\usepackage{fancyhdr}
\usepackage{setspace}

\newcites{supp}{References}

\begin{document}

\title{Self-supervised Augmentation Consistency \\for Adapting Semantic Segmentation}

\author{Nikita Araslanov$^1$ \hspace{1cm} Stefan Roth$^{1,2}$\\
$\ ^1$Department of Computer Science, TU Darmstadt \hspace{1cm} $\ ^2$ hessian.AI}

\maketitle
\thispagestyle{fancy}

\begin{abstract}

We propose an approach to domain adaptation for semantic segmentation that is both practical and highly accurate.
In contrast to previous work, we abandon the use of computationally involved adversarial objectives, network ensembles and style transfer.
Instead, we employ standard data augmentation techniques -- photometric noise, flipping and scaling -- and ensure consistency of the semantic predictions across these image transformations.
We develop this principle in a lightweight self-supervised framework trained on co-evolving pseudo labels without the need for cumbersome extra training rounds.
Simple in training from a practitioner's standpoint, our approach is remarkably effective.
We achieve significant improvements of the state-of-the-art segmentation accuracy after adaptation, consistent both across different choices of the backbone architecture and adaptation scenarios.
\blfootnote{
Code is available at \href{https://github.com/visinf/da-sac}{https://github.com/visinf/da-sac}.
}

\end{abstract}

\section{Introduction}
\label{sec:intro}

Unsupervised domain adaptation (UDA) is a variant of semi-supervised learning \cite{blum1998combining}, where the available unlabelled data comes from a different distribution than the annotated dataset \cite{Ben-DavidBCP06}.
A case in point is to exploit synthetic data, where annotation is more accessible compared to the costly labelling of real-world images \cite{RichterVRK16,RosSMVL16}.
Along with some success in addressing UDA for semantic segmentation \cite{TsaiHSS0C18,VuJBCP19,0001S20,ZouYKW18}, the developed methods are growing increasingly sophisticated and often combine style transfer networks, adversarial training or network ensembles \cite{KimB20a,LiYV19,TsaiSSC19,Yang_2020_ECCV}.
This increase in model complexity impedes reproducibility, potentially slowing further progress.

In this work, we propose a UDA framework reaching state-of-the-art segmentation accuracy (measured by the Intersection-over-Union, IoU) without incurring substantial training efforts.
Toward this goal, we adopt a simple semi-supervised approach, \emph{self-training} \cite{ChenWB11,lee2013pseudo,ZouYKW18}, used in recent works only in conjunction with adversarial training or network ensembles \cite{ChoiKK19,KimB20a,Mei_2020_ECCV,Wang_2020_ECCV,0001S20,Zheng_2020_IJCV,ZhengY20}.
By contrast, we use self-training \emph{standalone}.
Compared to previous self-training methods \cite{ChenLCCCZAS20,Li_2020_ECCV,subhani2020learning,ZouYKW18,ZouYLKW19}, our approach also sidesteps the inconvenience of multiple training rounds, as they often require expert intervention between consecutive rounds.
We train our model using co-evolving pseudo labels end-to-end without such need.

\begin{figure}[t]%
    \centering
    \def\svgwidth{\linewidth}
    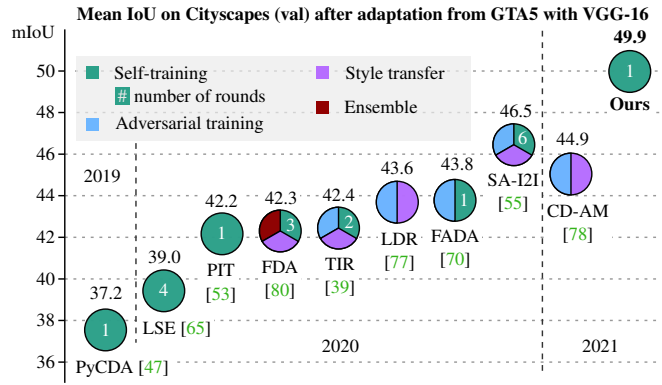
    \caption{\textbf{Results preview.} Unlike much recent work that combines multiple training paradigms, such as adversarial training and style transfer, our approach retains the modest single-round training complexity of self-training, yet improves the state of the art for adapting semantic segmentation by a significant margin.}
    \label{fig:preview}
\end{figure}

Our method leverages the ubiquitous \emph{data augmentation} techniques from fully supervised learning \cite{deeplabv3plus2018,ZhaoSQWJ17}: photometric jitter, flipping and multi-scale cropping.
We enforce \emph{consistency} of the semantic maps produced by the model across these image perturbations.
The following assumption formalises the key premise:

\myparagraph{Assumption 1.}
Let $f: \mathcal{I} \rightarrow \mathcal{M}$ represent a pixelwise mapping from images $\mathcal{I}$ to semantic output $\mathcal{M}$.
Denote $\rho_{\bm{\epsilon}}: \mathcal{I} \rightarrow \mathcal{I}$ a photometric image transform and, similarly, $\tau_{\bm{\epsilon}'}: \mathcal{I} \rightarrow \mathcal{I}$ a spatial similarity transformation, where $\bm{\epsilon},\bm{\epsilon}'\sim p(\cdot)$ are control variables following some pre-defined density (\eg, $p \equiv \mathcal{N}(0, 1)$).
Then, for any image $I \in \mathcal{I}$, $f$ is \emph{invariant} under $\rho_{\bm{\epsilon}}$ and \emph{equivariant} under $\tau_{\bm{\epsilon}'}$, \ie~$f(\rho_{\bm{\epsilon}}(I)) = f(I)$ and $f(\tau_{\bm{\epsilon}'}(I)) = \tau_{\bm{\epsilon}'}(f(I))$.

\smallskip
\noindent Next, we introduce a training framework using a \emph{momentum network} -- a slowly advancing copy of the original model.
The momentum network provides stable, yet recent targets for model updates, as opposed to the fixed supervision in model distillation \cite{Chen0G18,Zheng_2020_IJCV,ZhengY20}.
We also re-visit the problem of long-tail recognition in the context of generating pseudo labels for self-supervision.
In particular, we maintain an \emph{exponentially moving class prior} used to discount the confidence thresholds for those classes with few samples and increase their relative contribution to the training loss.
Our framework is simple to train, adds moderate computational overhead compared to a fully supervised setup, yet sets a new state of the art on established benchmarks (\cf \cref{fig:preview}).

\pagestyle{plain}
\section{Related Work}

Most of the work on scene adaptation for semantic segmentation has been influenced by a parallel stream of work on domain adaptation (DA) and semi-supervised learning for image classification \cite{FrenchMF18,ganin2016domain,GrandvaletB04,LiWSHL18,LongC0J18}.
The main idea behind these methods is to formulate an upper bound on the target risk using the so-called $\mathcal{H} \Delta \mathcal{H}$-divergence \cite{Ben-DavidBCKPV10}.
In a nutshell, it defines the discrepancy between the marginals of the source and target data by means of a binary classifier.
In the following, we briefly review implementation variants of this idea in the context of semantic segmentation.

\myparagraph{Learning domain-invariant representations.}

Adversarial feature alignment follows the GAN framework \cite{ganin2016domain,NIPS2014_5423} and minimises the gap between the source and target feature representations in terms of some distance (\eg, Wasserstein in \cite{LeeBBU19}).
The discriminator can be employed at multiple scales \cite{Chen0G18,TsaiHSS0C18,Yang_2020_ECCV} and use local spatial priors \cite{ZhangQYNL020};
it can be conditional \cite{HongWYY18} and class-specific \cite{DuTYFXZYZ19,Luo0GYY19},
or align the features of `hard' and `easy' target samples \cite{PanSRLK20}.
Often, self-supervised losses, such as entropy minimisation \cite{VuJBCP19}, or a `conservative loss' \cite{zhu2018penalizing} assist in this alignment.

The alternative to adversarial feature alignment are more interpretable constraints, such as feature priors \cite{luo2019significance}, bijective source-target association \cite{KangW0ZH20} or aligning the domains directly in the image space with style transfer \cite{CycleGAN2017} used either alone \cite{WuHLUGLD18} or, most commonly, jointly with adversarial feature alignment \cite{ChangWPC19,ChenL0H19,GongLCG19,MustoZ20,Yang_2021_WACV,YangLSS20,ZhangQY0M18}.
One issue with style translation is to ensure semantic consistency despite the changes in appearance.
To address this, Hoffman \etal~\cite{HoffmanTPZISED18} use semantic and cycle-consistency losses, while Yang \etal~\cite{Yang_2020_ECCV} reconstruct the original image from its label-space representation.

These methods tend to be computationally costly and challenging to train, since they require concurrent training of one or more independent networks, \eg~discriminators or style transfer networks.
Although Yang and Soatto \cite{0001S20} obviate the need for style networks by incorporating the phase of a Fourier-transformed target image into a source sample, multiple networks have to be trained, each with its own pre-defined phase band.

\myparagraph{Self-training on pseudo labels.}
As a more computationally lightweight approach, self-training seeks high-quality pseudo supervision coming in the form of class predictions with high confidence.
Our work belongs to this category.
Most of such previous methods pre-compute the labels `offline', used subsequently to update the model, and repeat this process for several rounds \cite{Li_2020_ECCV,subhani2020learning,ZouYKW18,ZouYLKW19}.
More recent frameworks following this strategy have a composite nature: they rely on adversarial (pre-)training \cite{ChenCCTWS17,DongCSLX20,ZhengY20}, style translation \cite{ChoiKK19,0001S20} or both \cite{Mei_2020_ECCV,LiYV19,KimB20a,Wang_2020_ECCV,WangYWFXHHS20}.

\begin{table}
\footnotesize
\begin{tabularx}{\linewidth}{@{}X|ccccc|c@{}}
\toprule
Features & \shortstack{PIT \\ \cite{LvLCL20}} & \shortstack{LDR \\ \cite{Yang_2020_ECCV}} & \shortstack{SA-I2I \\ \cite{MustoZ20}} & \shortstack{IAST \\ \cite{Mei_2020_ECCV}} & \shortstack{RPT \\ \cite{ZhangQYNL020}} & Ours \\
\midrule
Adversarial training & & \cmark & \cmark & \cmark & \cmark & \\
\midrule
1-round training & \cmark & \cmark & (6) & (3) & (3) & \cmark \\
SOTA-VGG & & \cmark & \cmark & & & \cmark \\
SOTA-ResNet & & & & \cmark & \cmark  & \cmark \\
\bottomrule
\end{tabularx}
\caption{\textbf{Relation to state of the art.} Previous work reaches the state of the art in terms of IoU either with VGG-16 (SOTA-VGG) or ResNet-101 (SOTA-ResNet). Our framework uses neither adversarial training nor multiple training rounds (given in parentheses), yet outperforms the state of the art consistently in both cases.}
\label{table:related_work}
\end{table}

Training on co-evolving pseudo labels can be computationally unstable, hence requires additional regularisation.
Chen \etal~\cite{0001XC19} minimise the entropy with improved behaviour of the gradient near the saturation points.
Using fixed representations, be it from a `frozen' network \cite{Chen0G18,ZhengY20}, a fixed set of global \cite{LvLCL20} or self-generated local labels \cite{LianDLG19,TsaiSSC19,ZhangDG17}, further improves training robustness.

Overconfident predictions \cite{GuoPSW17} have direct consequences for the quality of pseudo labels.
Zou \etal~\cite{ZouYLKW19} attain some degree of confidence calibration via regularising the loss with prediction smoothing akin to temperature scaling \cite{GuoPSW17}.
Averaging the predictions of two classifiers \cite{Zheng_2020_IJCV}, or using Dropout-based sampling \cite{Cai_2020_CVPR,zhou2020uncertainty}, achieves the same goal.

\begin{figure*}[t]%
\subcaptionbox{\scriptsize Framework overview\label{fig:model_overview}}{%
    \def\svgwidth{0.44\linewidth}
    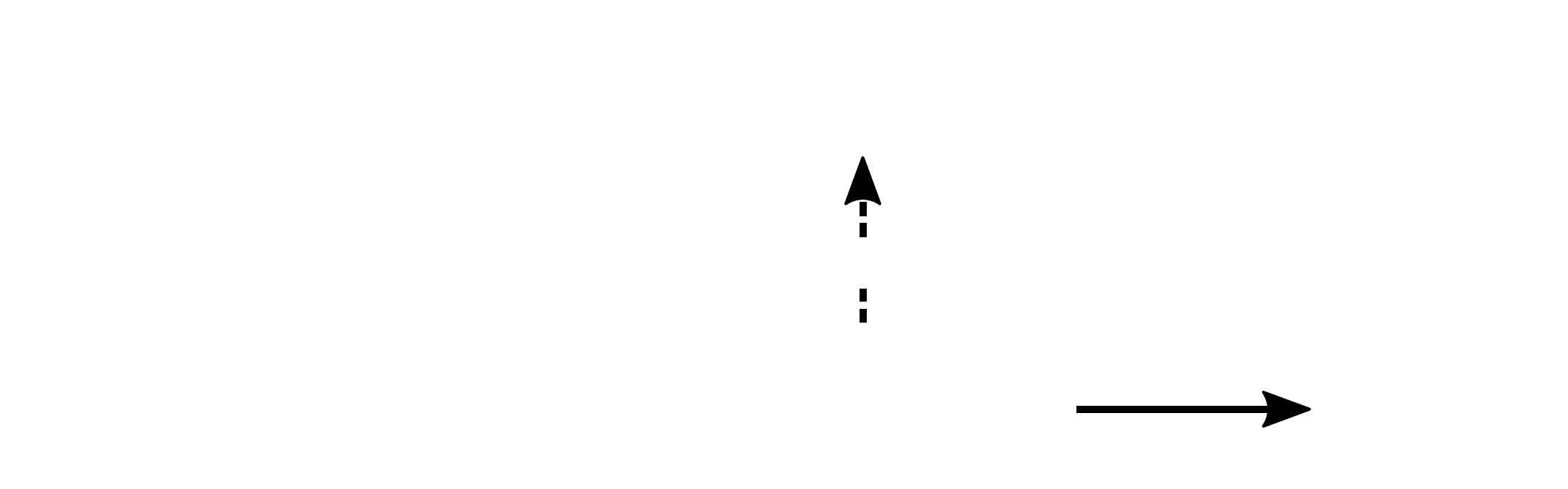
}\hfill
\subcaptionbox{\scriptsize Multi-scale crops and flips\label{fig:batch}}{%
    \def\svgwidth{0.26\linewidth}
\begingroup%
  \makeatletter%
  \providecommand\color[2][]{%
    \errmessage{(Inkscape) Color is used for the text in Inkscape, but the package 'color.sty' is not loaded}%
    \renewcommand\color[2][]{}%
  }%
  \providecommand\transparent[1]{%
    \errmessage{(Inkscape) Transparency is used (non-zero) for the text in Inkscape, but the package 'transparent.sty' is not loaded}%
    \renewcommand\transparent[1]{}%
  }%
  \providecommand\rotatebox[2]{#2}%
  \newcommand*\fsize{\dimexpr\f@size pt\relax}%
  \newcommand*\lineheight[1]{\fontsize{\fsize}{#1\fsize}\selectfont}%
  \ifx\svgwidth\undefined%
    \setlength{\unitlength}{228.24866047bp}%
    \ifx\svgscale\undefined%
      \relax%
    \else%
      \setlength{\unitlength}{\unitlength * \real{\svgscale}}%
    \fi%
  \else%
    \setlength{\unitlength}{\svgwidth}%
  \fi%
  \global\let\svgwidth\undefined%
  \global\let\svgscale\undefined%
  \makeatother%
  \begin{picture}(1,0.52532922)%
    \scriptsize
    \lineheight{1}%
    \setlength\tabcolsep{0pt}%
    \put(0,0){\includegraphics[width=\unitlength,page=1]{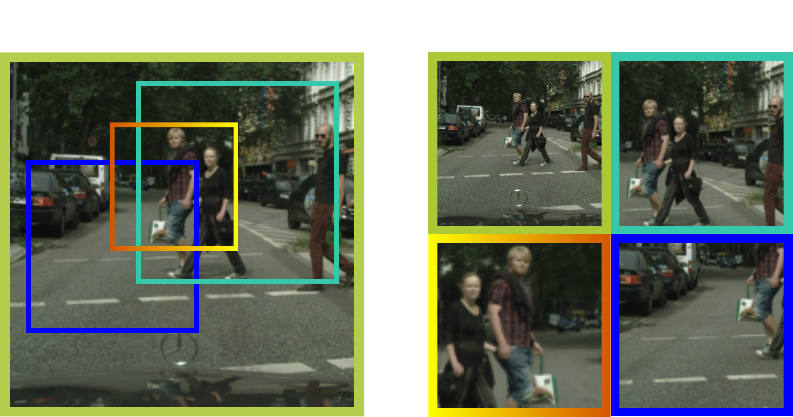}}%
    \put(0.08,0.485){\color[rgb]{0,0,0}\makebox(0,0)[lt]{\lineheight{1.25}\smash{\begin{tabular}[t]{l}Input Sample\end{tabular}}}}%
    \put(0.63292981,0.485){\color[rgb]{0,0,0}\makebox(0,0)[lt]{\lineheight{1.25}\smash{\begin{tabular}[t]{l}Input Batch\end{tabular}}}}%
    \put(0,0){\includegraphics[width=\unitlength,page=2]{figures/batch/batch.pdf}}%
  \end{picture}%
\endgroup%

}\hfill
\subcaptionbox{\scriptsize Multi-scale fusion\label{fig:fusion}}{%
    \def\svgwidth{0.26\linewidth}
\begingroup%
  \makeatletter%
  \providecommand\color[2][]{%
    \errmessage{(Inkscape) Color is used for the text in Inkscape, but the package 'color.sty' is not loaded}%
    \renewcommand\color[2][]{}%
  }%
  \providecommand\transparent[1]{%
    \errmessage{(Inkscape) Transparency is used (non-zero) for the text in Inkscape, but the package 'transparent.sty' is not loaded}%
    \renewcommand\transparent[1]{}%
  }%
  \providecommand\rotatebox[2]{#2}%
  \newcommand*\fsize{\dimexpr\f@size pt\relax}%
  \newcommand*\lineheight[1]{\fontsize{\fsize}{#1\fsize}\selectfont}%
  \ifx\svgwidth\undefined%
    \setlength{\unitlength}{228.2483631bp}%
    \ifx\svgscale\undefined%
      \relax%
    \else%
      \setlength{\unitlength}{\unitlength * \real{\svgscale}}%
    \fi%
  \else%
    \setlength{\unitlength}{\svgwidth}%
  \fi%
  \global\let\svgwidth\undefined%
  \global\let\svgscale\undefined%
  \makeatother%
  \begin{picture}(1,0.52376449)%
    \scriptsize
    \lineheight{1}%
    \setlength\tabcolsep{0pt}%
    \put(0,0){\includegraphics[width=\unitlength,page=1]{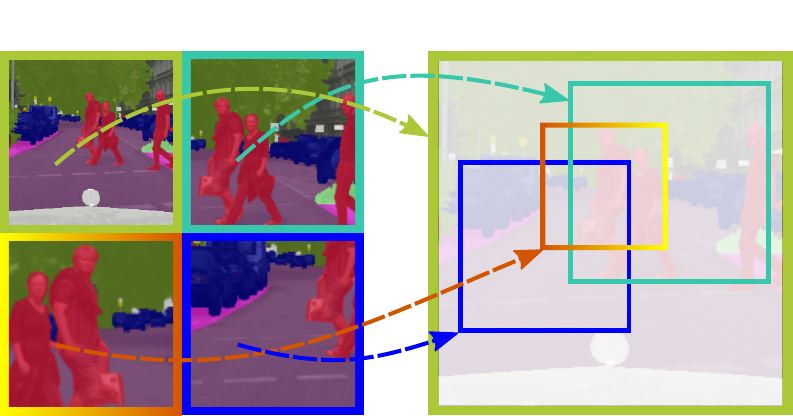}}%
    \put(0.07,0.485){\color[rgb]{0,0,0}\makebox(0,0)[lt]{\lineheight{1.25}\smash{\begin{tabular}[t]{l}Output Masks\end{tabular}}}}%
    \put(0.576,0.485){\color[rgb]{0,0,0}\makebox(0,0)[lt]{\lineheight{1.25}\smash{\begin{tabular}[t]{l}Fused Prediction\end{tabular}}}}%
  \end{picture}%
\endgroup%

}
\caption{\textbf{Overview.} The segmentation network in our framework \emph{\subref{fig:model_overview}} maintains a slow copy of itself, the momentum network, which provides stable targets for self-supervision. In addition to encouraging semantic invariance \wrt the photometric noise, we facilitate consistent predictions across multiple scales and flips by first \emph{\subref{fig:batch}} feeding random multi-scale crops and flips to the momentum network and then \emph{\subref{fig:fusion}} fusing the predictions by simple averaging to produce the pseudo-supervision targets.} %
\label{fig:batch_fusion}%
\vspace{-0.5em}
\end{figure*}

\myparagraph{Spatial priors.}
Different from DA for classification, the characteristic feature of adaptation methods for segmentation is the use of spatial priors.
Local priors have been enforced patch-wise \cite{Chen0G18,LianDLG19,TsaiSSC19} and in the form of pre-computed super-pixels \cite{ZhangDG17,ZhangQYNL020}.
Although global spatial priors have also been used \cite{ZouYKW18}, their success hinges on the similarity of the semantic layout in the current benchmarks.

\myparagraph{Relation to our approach.}
As shown in Table~\ref{table:related_work}, our work streamlines the training process.
First, we do not use adversarial training, as feature invariance alone does not guarantee \textit{label} invariance \cite{JohanssonSR19,0002CZG19}.
Second, we train our model with co-evolving pseudo labels in one round.
Our framework bears resemblance to the noisy mean teacher \cite{XieLHL20} and combines consistency regularisation \cite{BachmanAP14,SajjadiJT16,SohnBCZZRCKL20,XieDHL020} with self-ensembling \cite{LaineA17,tarvainen2017mean}.
Similar approaches have been explored in medical imaging \cite{LiY0FH18,PeroneBBC19} and concurrent UDA work \cite{Wang_Yang_Betke_2021}, albeit limited in the scope of admissible augmentations.
We leverage photometric invariance, scale and flip equivariance \cite{WangZKSC20} to extract high-fidelity pseudo supervision instead of more computationally expensive sampling techniques \cite{KendallG17}.
Contrary to \cite{subhani2020learning}, we find that scale alone is not predictive of the label quality, hence we average the predictions produced at multiple scales and flips.
This parallels uncertainty estimation using test-time augmentation \cite{ayhan2018test}, but at training time \cite{BerthelotCGPOR19}.

\section{Self-Supervised Augmentation Consistency}

\subsection{Framework overview}

Shown in \cref{fig:model_overview}, our framework comprises a segmentation network, which we intend to adapt to a target domain, and its slowly changing copy updated with a momentum, a \emph{momentum network}.
To perform self-supervised scene adaptation, we first supply a batch of random crops and horizontal flips from a sample image of the target domain to both networks.
For each pixel we average the predictions (\ie~semantic masks) from the momentum network after the appropriate inverse spatial transformation.
We then create a pseudo ground truth by selecting confident pixels from the averaged map using thresholds based on running statistics, which are capable of adapting to individual samples.
Finally, the segmentation network uses stochastic gradient descent to update its parameters \wrt these pseudo labels.

Our approach closely resembles the mean teacher framework \cite{FrenchMF18,tarvainen2017mean} and temporal ensembling \cite{IzmailovPGVW18,LaineA17}.
However, as we will show empirically, the ensembling property itself plays only an auxiliary role.
More importantly, akin to the critic network in reinforcement learning \cite{LillicrapHPHETS15} and the momentum encoder in unsupervised learning \cite{He0WXG20}, our momentum network provides stable targets for self-supervised training of the segmentation network.
This view allows us to focus on the target-generating process, detailed next.

\subsection{Batch construction}
\label{sec:batch}

For each sampled target image, we generate $N$ crops with random scales, flips and locations, but preserving the aspect ratio.
We re-scale the crops as well as the original image to a fixed input resolution $h \times w$
and pass them as the input to the networks.
\cref{fig:batch} demonstrates this process.
Following the noisy student model in image classification \cite{XieLHL20}, the input to the segmentation network additionally
undergoes a photometric augmentation:
we add random colour jitter and smooth the images with a Gaussian filter at random.
The momentum network, on the other hand, receives a `clean' input, \ie~without such augmentations.
This is to encourage model invariance to photometric perturbations.

\subsection{Self-supervision}
\label{sec:selftrain}

\paragraph{Multi-scale fusion.}
We re-project the output masks from the momentum network back to the original image canvas of size $h \times w$, as illustrated in \cref{fig:fusion}.
For each pixel, the overlapping areas average their predictions.
Note that some pixels may lie outside the crops, hence contain the result of a single forward pass with the original image.
We keep these predictions intact.
The merged maps are then used to extract the pseudo masks for self-supervision.

\myparagraph{A short long-tail interlude.}
Handling rare classes (\ie\ classes with only a few training samples) is notoriously difficult in recognition \cite{GuptaDG19}.
For semantic segmentation, we here distinguish between the classes with low image-level (\eg, ``truck'', ``bus'') and pixel-level (\eg, ``traffic light'', ``pole'') frequency.
While generating self-supervision, we take special care of these cases and encourage
\begin{enumerate*}[label=(\roman*), font=\itshape]
  \item lower thresholds for selecting their pseudo labels,
  \item increased contributions to the gradient with a focal loss, and
  \item employ importance sampling.
\end{enumerate*}
We describe these in detail next.

\begin{figure}[t]
\input{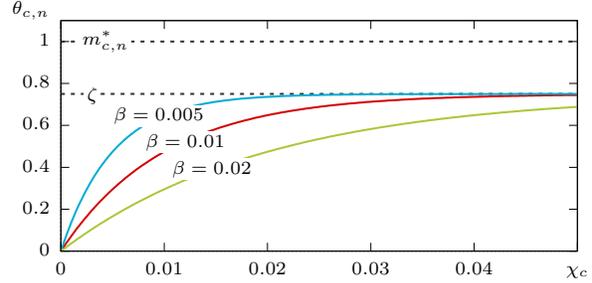}
\caption{\textbf{Sample-based moving threshold.} Our thresholding scheme has two hyperparameters, $\zeta$ and $\beta$. In this example, $m^\ast_{c,n} = 1$ and $\zeta=0.75$. Predominant classes (\eg, ``road'') have $\chi_c \gg 0$, hence their threshold approximates $\zeta m^\ast_{c,n}$. Long-tail classes (\eg, ``traffic light'') have $\chi_c \approx 0$ and their thresholds are further reduced with a steepness controlled by $\beta$ (see Eq.~\ref{eq:sbrt}).}
\label{fig:threshold}
\end{figure}

\myparagraph{Sample-based moving threshold.}
Most previous work with self-training employs multi-round training that requires interrupting the training process and re-generating the pseudo labels \cite{Li_2020_ECCV,LiYV19,Mei_2020_ECCV,subhani2020learning,ZouYKW18}.
One of the reasons is the need to re-compute the thresholds for filtering the pseudo labels for supervision, which requires traversing the predictions for the complete target dataset with the model parameters fixed.
In pursuit of our goal of enabling end-to-end training without expert intervention, we take a different approach and compute the thresholds on-the-go.
As the main ingredient, we maintain an \emph{exponentially moving class prior}.
In detail, for each softmax prediction of the momentum network, we first compute a prior estimate of the probability that a pixel in sample $n$ belongs to class $c$ as
\begin{equation}
\chi_{c,n} = \frac{1}{h w} \sum_{i,j} m_{c,n,i,j},
\label{eq:prior}
\end{equation}
where $m_{c,n,:,:}$ is the mask prediction for class $c$ (with resolution $h \times w$).
We keep an exponentially moving average after each training iteration $t$ with a momentum $\gamma_\chi \in [0, 1]$:
\begin{equation}
\chi_c^{t+1} = \gamma_\chi \chi^t_c + (1 - \gamma_\chi) \chi_{c,n}.
\end{equation}
Our sample-based moving threshold $\theta_{c,n}$ takes lower values when the moving prior $\chi_c \approx 0$ (\ie~for long-tail classes), but is bounded from above as $\chi_c \rightarrow 1$.
We define it as
\begin{equation}
\theta_{c,n} = \zeta \big(1 - e^{-\chi_c / \beta}\big) m^\ast_{c,n},
\label{eq:sbrt}
\end{equation}
where $\beta$ and $\zeta$ are hyperparameters and $m^\ast_{c,n}$ is the predicted peak confidence score for class $c$, \ie
\begin{equation}
m^\ast_{c,n} = \max_{i,j} m_{c,n,i,j}.
\label{eq:peak-class-confidence}
\end{equation}
\cref{fig:threshold} plots \cref{eq:sbrt} as a function of the moving class prior $\chi_c$ for a selection of $\beta$.
For predominant classes (\eg, ``road''), the exponential term has nearly no effect; the threshold is static \wrt the peak class confidence, \ie $\theta_{c,n} \approx \zeta m^\ast_{c,n}$.
However, for long-tail classes such that $\chi_c \approx \beta$, the threshold is lower than this upper bound, hence more pixels for these classes are selected for supervision.
To obtain the pseudo labels, we apply the threshold $\theta_{c,n}$ to the peak predictions of the merged output from the momentum network:
\begin{equation}
\hat{m}_{n,i,j} =
\begin{cases}
      c^\ast & m_{c^\ast,n,i,j} > \theta_{c,n} \\
      \text{ignore} & \text{otherwise},
\end{cases}
\label{eq:pseudo}
\end{equation}
where $c^\ast = \argmax_{c} m_{c,n,i,j}$ is the dominant class for that pixel.
Note that the pixels with confidence values lower than the threshold, as well as non-dominant predictions, will be ignored in the self-supervised loss.

\myparagraph{Focal loss with confidence regularisation.}
Our loss function incorporates a focal multiplier \cite{LinGGHD20} to further increase the contribution of the long-tail classes in the gradient signal.
Unlike previous work \cite{LinGGHD20,subhani2020learning}, however, our moving class prior $\chi_c$ regulates the focal term:
\begin{equation}
\mathcal{L}^t_{n}(\bar{m}, m \mid \phi) = -m_{c^\ast\!,n} (1 - \chi_{c^\ast})^\lambda \log(\bar{m}_{c^\ast\!,n}),
\label{eq:loss}
\end{equation}
where $\bar{m}$ is the prediction of the segmentation network with parameters $\phi$, the pseudo label $c^\ast$ derives from $\hat{m}$ in \cref{eq:pseudo} and $\lambda$ is a hyperparameter of the focal term.
Recall that low values of $\chi_c$ signify a long-tail category, hence should have a higher weight.
High values of $\lambda$ (\ie~$> 1$) increase the relative weighting on the long-tail classes, while setting $\lambda = 0$ disables the focal term.
Note that we also regularise our loss with the confidence value of the momentum network, $m_{c^\ast,n}$ (\Eq~\ref{eq:peak-class-confidence}).
In case of an incorrect pseudo label, we expect this confidence to be low and to regularise the training owing to its calibration with the multi-scale fusion.
We minimise the loss in \cref{eq:loss}, applied for each pixel, \wrt $\phi$.

\begin{figure*}[t]
\begin{subfigure}{.2\linewidth}
  \centering
    \includegraphics[width=0.98\linewidth]{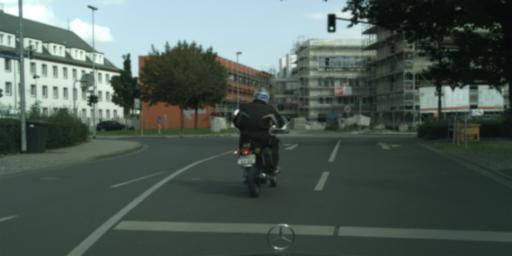}\\
    \includegraphics[width=0.98\linewidth]{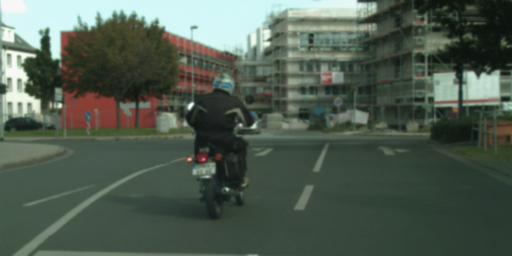}\\
    \includegraphics[width=0.98\linewidth]{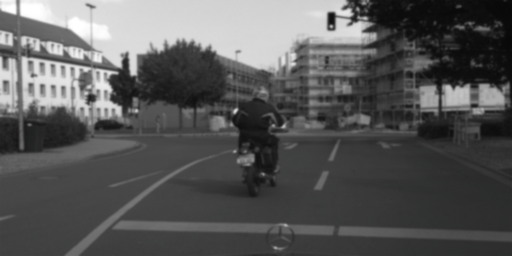}\\
    \includegraphics[width=0.98\linewidth]{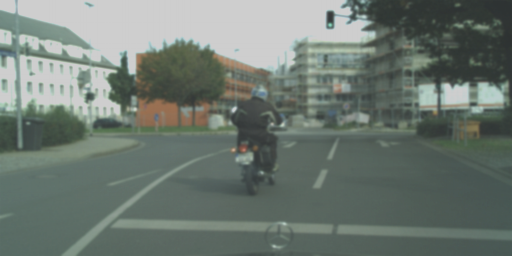}
  \caption{\scriptsize Input}
  \label{fig:ex_fusion_input}
\end{subfigure}%
\begin{subfigure}{.2\linewidth}
  \centering
    \includegraphics[width=0.98\linewidth]{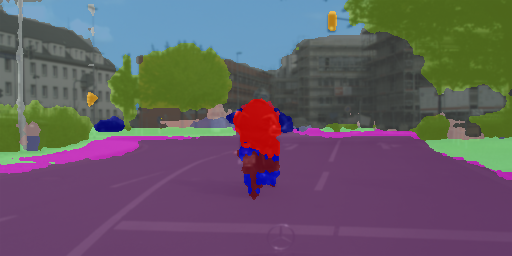}\\
    \includegraphics[width=0.98\linewidth]{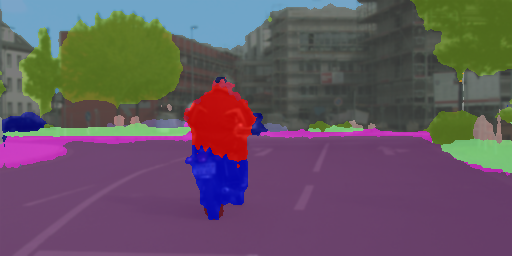}\\
    \includegraphics[width=0.98\linewidth]{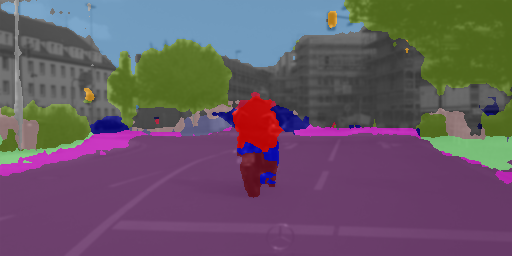}\\
    \includegraphics[width=0.98\linewidth]{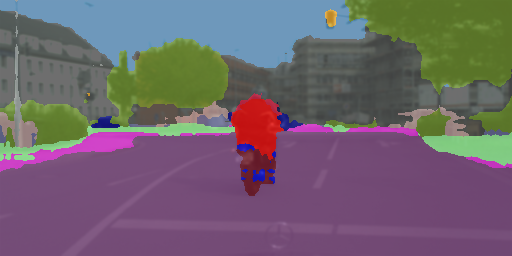}
  \caption{\scriptsize Segmentation net output}
  \label{fig:ex_fusion_seg}
\end{subfigure}%
\begin{subfigure}{.2\linewidth}
  \centering
    \includegraphics[width=0.98\linewidth]{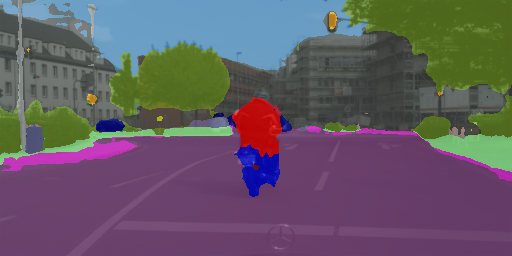}\\
    \includegraphics[width=0.98\linewidth]{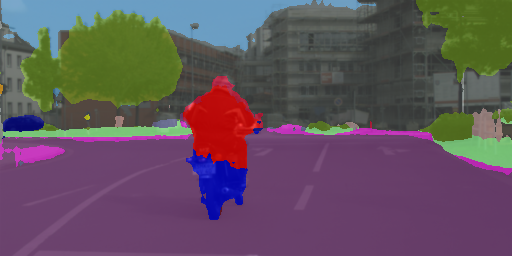}\\
    \includegraphics[width=0.98\linewidth]{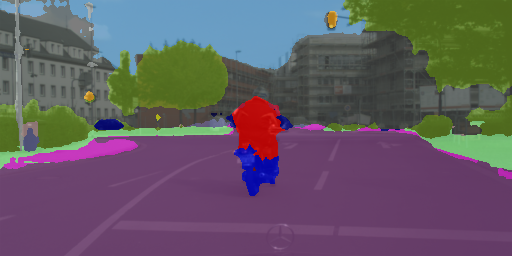}\\
    \includegraphics[width=0.98\linewidth]{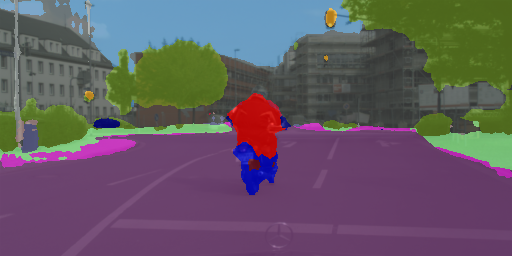}
  \caption{\scriptsize Momentum net output}
  \label{fig:ex_fusion_mom}
\end{subfigure}%
\begin{subfigure}{.2\linewidth}
  \centering
    \includegraphics[width=0.98\linewidth]{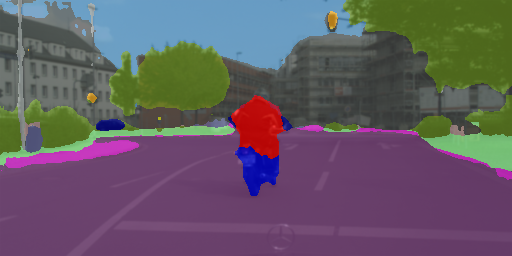}\\
    \includegraphics[width=0.98\linewidth]{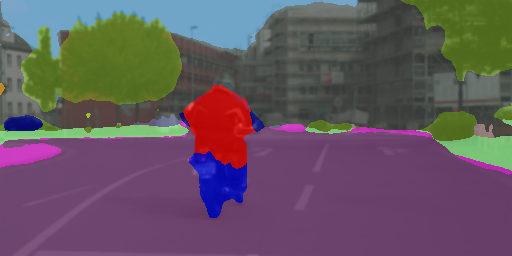}\\
    \includegraphics[width=0.98\linewidth]{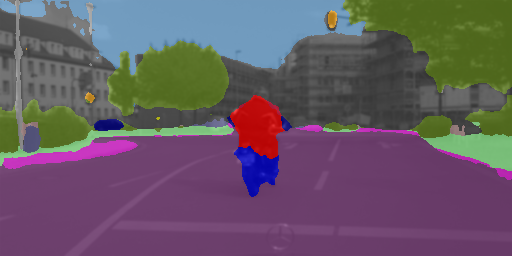}\\
    \includegraphics[width=0.98\linewidth]{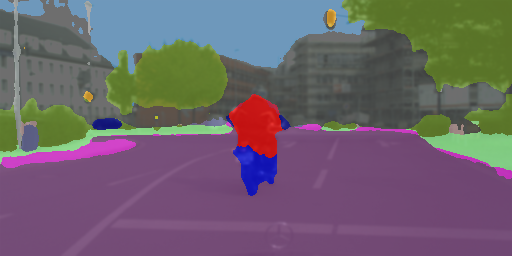}
  \caption{\scriptsize Fused prediction}
  \label{fig:ex_fusion_merged}
\end{subfigure}%
\begin{subfigure}{.2\linewidth}
  \centering
    \includegraphics[width=0.98\linewidth]{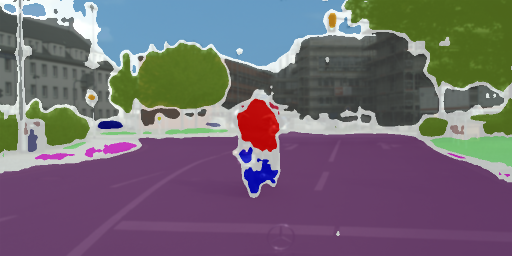}\\
    \includegraphics[width=0.98\linewidth]{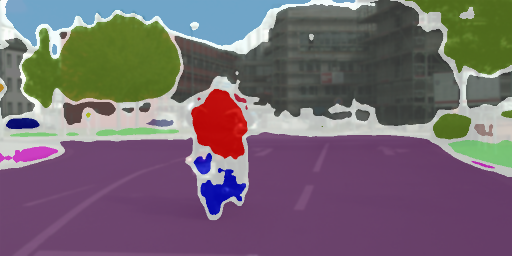}\\
    \includegraphics[width=0.98\linewidth]{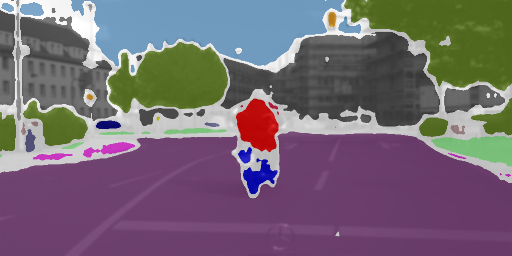}\\
    \includegraphics[width=0.98\linewidth]{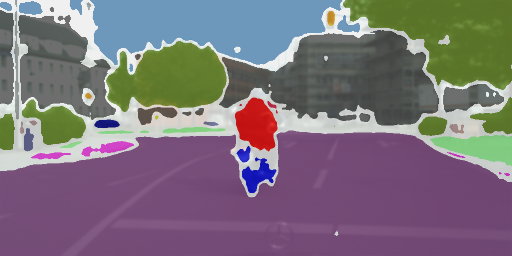}
  \caption{\scriptsize Pseudo labels}
  \label{fig:ex_fusion_pseudo}
\end{subfigure}
\caption{\textbf{Self-supervision example.} In this image sample \emph{\subref{fig:ex_fusion_input}} and its crops, the segmentation network \emph{\subref{fig:ex_fusion_seg}} tends to mistake the ``motorcycle'' for a ``bicycle''.
The momentum network \emph{\subref{fig:ex_fusion_mom}} improves on this prediction, but may still produce an inconsistent labelling.
Averaging the predictions over multiple scales \emph{\subref{fig:ex_fusion_merged}} corrects this inconsistency, allowing to produce high-precision pseudo labels \emph{\subref{fig:ex_fusion_pseudo}} for self-supervision.}
\label{fig:example_fusion}
\vspace{-0.6em}
\end{figure*}

\subsection{Training}
\label{sec:training}

\paragraph{Pre-training with source-only loss.}
Following \cite{LianDLG19,ZhangQYNL020}, we use Adaptive Batch Normalisation (ABN) \cite{LiWSHL18} to jump-start our model on the segmentation task by minimising the cross-entropy loss on the source data only.
In our experiments, we found it unnecessary to re-compute the mean and the standard deviation only at the end of the training.
Instead, in pre-training we alternate batches of source and target images, but ignore the loss for the latter.
For a target batch, this implies updating the running mean and the standard deviation in the Batch Normalisation (BN) \cite{IoffeS15} layers and leaving the remaining model parameters untouched.

\myparagraph{Importance sampling.}
Our loss function in \cref{eq:loss} accounts for long-tail classes with a high image frequency (\eg, ``traffic light'', ``pole''), and may not be effective for the classes appearing in only few samples (\eg, ``bus'', ``train'').
To alleviate this imbalance, we use importance sampling \cite{DoucetFG01} and increase the sample frequency of these long-tail classes.
We minimise the \emph{expected} target loss by re-sampling the target images using the density $p_t$:
\begin{equation}
\underset{\phi}{\min} \: \mathbb{E}_{n \sim p_t} \big[\mathcal{L}^t_n (\phi) \big].
\label{eq:is_obj}
\end{equation}
To obtain $p_t$, we use our pre-trained segmentation network and pre-compute $\chi_{c,n}$, the class prior estimate, for each image $n$ using \cref{eq:prior}.
At training time, we \textit{(i)} sample a semantic class $c$ uniformly, and then \textit{(ii)} obtain a target sample $l$ with probability 
\begin{equation}
\hat{\chi}_{c,l} = \frac{\chi_{c,l}}{\sum_n \chi_{c,n}}.
\end{equation}
This two-step sampling process ensures that all images have non-zero sample probability owing to the prevalent classes for which $\hat{\chi}_{c,l} > 0$ for all $l$ (\eg, ``road'' in urban scenes).

\myparagraph{Joint target-source training.}
We train the segmentation network with stochastic gradient descent using the cross-entropy loss for the source and our focal loss for the target data sampled from $p_t$, as defined by \cref{eq:loss,eq:is_obj}.
\cref{fig:example_fusion} illustrates the synthesis of pseudo labels.
We periodically update the parameters $\psi$ of the momentum network as
\begin{equation}
\psi_{t+1} = \gamma_\psi \psi_t + (1 - \gamma_\psi) \phi,
\end{equation}
where $\phi$ are the parameters of the segmentation network.
$\gamma_\psi$ regulates the pace of the updates: low values result in faster, but unstable training,
while high $\gamma_\psi$ leads to a premature and suboptimal convergence.
We keep $\gamma_\psi$ moderate, but update the momentum network only every $T$ iterations.

\section{Experiments}
\label{sec:exp}

\paragraph{Datasets.}
In our experiments we use three datasets.
The Cityscapes dataset \cite{CordtsORREBFRS16} contains $\text{2048} \times \text{1024}$ images from real-world traffic scenes, split into \num{2975} images for training and \num{500} for validation.
The GTA5 dataset \cite{RichterVRK16} contains \num{24966} synthetic scenes with resolution $\text{1914} \times \text{1052}$ and pixelwise annotation aided by the GTA5 game engine.
We also use the SYNTHIA-RAND-CITYSCAPES subset of the SYNTHIA dataset \cite{RosSMVL16}, which contains \num{9400} synthetic images with resolution $\text{1280} \times \text{760}$ and a semantic annotation compatible with Cityscapes.

\begin{table*}
\footnotesize
\begin{tabularx}{\linewidth}{@{}X|S[table-format=2.1]@{\hspace{0.7em}}S[table-format=2.1]@{\hspace{0.7em}}S[table-format=2.1]@{\hspace{0.7em}}S[table-format=2.1]@{\hspace{0.7em}}S[table-format=2.1]@{\hspace{0.7em}}S[table-format=2.1]@{\hspace{0.7em}}S[table-format=2.1]@{\hspace{0.7em}}S[table-format=2.1]@{\hspace{0.7em}}S[table-format=2.1]@{\hspace{0.7em}}S[table-format=2.1]@{\hspace{0.7em}}S[table-format=2.1]@{\hspace{0.7em}}S[table-format=2.1]@{\hspace{0.7em}}S[table-format=2.1]@{\hspace{0.7em}}S[table-format=2.1]@{\hspace{0.7em}}S[table-format=2.1]@{\hspace{0.7em}}S[table-format=2.1]@{\hspace{0.7em}}S[table-format=2.1]@{\hspace{0.7em}}S[table-format=2.1]@{\hspace{0.7em}}S[table-format=2.1]@{\hspace{0.7em}}|S[table-format=2.1]@{}}
\toprule
Method & {road} & {sidew} & {build} & {wall} & {fence} & {pole} & {light} & {sign} & {veg} & {terr} & {sky} & {pers} & {ride} & {car} & {truck} & {bus} & {train} & {moto} & {bicy} & {mIoU} \\
\midrule
\multicolumn{21}{@{}l}{\scriptsize \textit{Backbone: VGG-16}} \\
\midrule
CyCADA \cite{HoffmanTPZISED18} & 85.2 & 37.2 & 76.5 & 21.8 & 15.0 & 23.8 & 22.9 & 21.5 & 80.5 & 31.3 & 60.7 & 50.5 & 9.0 & 76.9 & 17.1 & 28.2 & 4.5 & 9.8 & 0.0 & 35.4 \\
ADVENT \cite{VuJBCP19} & 86.9 & 28.7 & 78.7 & 28.5 & 25.2 & 17.1 & 20.3 & 10.9 & 80.0 & 26.4 & 70.2 & 47.1 & 8.4 & 81.5 & 26.0 & 17.2 & 18.9 & 11.7 & 1.6 & 36.1 \\
CBST \cite{ZouYKW18} & 90.4 & 50.8 & 72.0 & 18.3 & 9.5 & 27.2 & 8.6 & 14.1 & 82.4 & 25.1 & 70.8 & 42.6 & 14.5 & 76.9 & 5.9 & 12.5 & 1.2 & 14.0 & 28.6 & 36.1 \\
PyCDA \cite{LianDLG19} & 86.7 & 24.8 & 80.9 & 21.4 & 27.3 & 30.2 & 26.6 & 21.1 & \bfseries 86.6 & 28.9 & 58.8 & 53.2 & 17.9 & 80.4 & 18.8 & 22.4 & 4.1 & 9.7 & 6.2 & 37.2 \\
PIT \cite{LvLCL20} & 86.2 & 35.0 & 82.1 & 31.1 & 22.1 & 23.2 & 29.4 & 28.5 & 79.3 & 31.8 & 81.9 & 52.1 & 23.2 & 80.4 & 29.5 & 26.9 & \bfseries 30.7 & 20.5 & 1.2 & 41.8 \\
FDA \cite{0001S20} & 86.1 & 35.1 & 80.6 & 30.8 & 20.4 & 27.5 & 30.0 & 26.0 & 82.1 & 30.3 & 73.6 & 52.5 & 21.7 & 81.7 & 24.0 & 30.5 & 29.9 & 14.6 & 24.0 & 42.2 \\
LDR \cite{Yang_2020_ECCV} & 90.1 & 41.2 & 82.2 & 30.3 & 21.3 & 18.3 & 33.5 & 23.0 & 84.1 & 37.5 & 81.4 & 54.2 & 24.3 & 83.0 & 27.6 & 32.0 & 8.1 & \bfseries 29.7 & 26.9 & 43.6 \\
FADA \cite{Wang_2020_ECCV} & \bfseries 92.3 & 51.1 & 83.7 & 33.1 & 29.1 & 28.5 & 28.0 & 21.0 & 82.6 & 32.6 & 85.3 & 55.2 & 28.8 & 83.5 & 24.4 & 37.4 & 0.0 & 21.1 & 15.2 & 43.8 \\
CD-AM \cite{Yang_2021_WACV} & 90.1 & 46.7 & 82.7 & \bfseries 34.2 & 25.3 & 21.3 & 33.0 & 22.0 & 84.4 & \bfseries 41.4 & 78.9 & 55.5 & 25.8 & 83.1 & 24.9 & 31.4 & 20.6 & 25.2 & 27.8 & 44.9 \\
SA-I2I \cite{MustoZ20} & 91.1 & 46.4 & 82.9 & 33.2 & 27.9 & 20.6 & 29.0 & 28.2 & 84.5 & 40.9 & 82.3 & 52.4 & 24.4 & 81.2 & 21.8 & 44.8 & 31.5 & 26.5 & \bfseries 33.7 & 46.5 \\
\midrule
Baseline (ours) & 81.5 & 28.6 & 79.5 & 23.2 & 21.1 & 31.3 & 28.2 & 18.5 & 75.6 & 14.9 & 72.2 & 58.0 & 17.1 & 81.1 & 19.7 & 26.3 & 13.7 & 12.9 & 2.1 & 37.1 \\
SAC (ours) & 90.0 & \bfseries 53.1 & \bfseries 86.2 & 33.8 & \bfseries 32.7 & \bfseries 38.2 & \bfseries 46.0 & \bfseries 40.3 & 84.2 & 26.4 & \bfseries 88.4 & \bfseries 65.8 & \bfseries 28.0 & \bfseries 85.6 & \bfseries 40.6 & \bfseries 52.9 & 17.3 & 13.7 & 23.8 & \bfseries 49.9 \\
\midrule
\multicolumn{21}{@{}l}{\scriptsize \textit{Backbone: ResNet-101}} \\
\midrule
PyCDA$^\dagger$ \cite{LianDLG19} & 90.5 & 36.3 & 84.4 & 32.4& 28.7 & 34.6 & 36.4 & 31.5 & 86.8 & 37.9 & 78.5 & 62.3 & 21.5 & 85.6 & 27.9 & 34.8 & 18.0 & 22.9 & 49.3 & 47.4 \\
CD-AM \cite{Yang_2021_WACV} & 91.3 & 46.0 & 84.5 & 34.4 & 29.7 & 32.6 & 35.8 & 36.4 & 84.5 & 43.2 & 83.0 & 60.0 & 32.2 & 83.2 & 35.0 & 46.7 & 0.0 & 33.7 & 42.2 & 49.2 \\
FADA \cite{Wang_2020_ECCV} & 92.5 & 47.5 & 85.1 & 37.6 & \bfseries 32.8 & 33.4 & 33.8 & 18.4 & 85.3 & 37.7 & 83.5 & 63.2 & \bfseries 39.7 & 87.5 & 32.9 & 47.8 & 1.6 & 34.9 & 39.5 & 49.2 \\
LDR \cite{Yang_2020_ECCV} & 90.8 & 41.4 & 84.7 & 35.1 & 27.5 & 31.2 & 38.0 & 32.8 & 85.6 & 42.1 & 84.9 & 59.6 & 34.4 & 85.0 & 42.8 & 52.7 & 3.4 & 30.9 & 38.1 & 49.5 \\
FDA \cite{0001S20} & 92.5 & 53.3 & 82.4 & 26.5 & 27.6 & 36.4 & 40.6 & 38.9 & 82.3 & 39.8 & 78.0 & 62.6 & 34.4 & 84.9 & 34.1 & 53.1 & 16.9 & 27.7 & 46.4 &  50.5 \\
SA-I2I \cite{MustoZ20} & 91.2 & 43.3 & 85.2 & 38.6 & 25.9 & 34.7 & 41.3 & 41.0 & 85.5 & \bfseries 46.0 & 86.5 & 61.7 & 33.8 & 85.5 & 34.4 & 48.7 & 0.0 & 36.1 & 37.8 & 50.4 \\
PIT \cite{LvLCL20} & 87.5  & 43.4  & 78.8  & 31.2  & 30.2  & 36.3  & 39.9  & 42.0  & 79.2  & 37.1  & 79.3  & 65.4  & 37.5  & 83.2  & 46.0  & 45.6  & 25.7  & 23.5  & 49.9 & 50.6 \\
IAST \cite{Mei_2020_ECCV} & \bfseries 93.8 & \bfseries 57.8 & 85.1 & 39.5 & 26.7 & 26.2 & 43.1 & 34.7 & 84.9 & 32.9 & \bfseries 88.0 & 62.6 & 29.0 & 87.3 & 39.2 & 49.6 & 23.2 & 34.7 & 39.6 & 51.5 \\
RPT$^\dagger$ \cite{ZhangQYNL020} & 89.2 & 43.3 & 86.1 & 39.5 & 29.9 & 40.2 & \bfseries 49.6 & 33.1 & \bfseries 87.4 & 38.5 & 86.0 & 64.4 & 25.1 & \bfseries 88.5 & 36.6 & 45.8 & \bfseries 23.9 & \bfseries 36.5 & \bfseries 56.8 &  52.6 \\
\midrule
Baseline (ours) & 80.2 & 29.3 & 76.8 & 23.8 & 21.9 & 37.7 & 35.4 & 21.1 & 79.8 & 21.3 & 75.0 & 59.5 & 17.5 & 83.5 & 22.4 & 33.4 & 13.0 & 30.7 & 12.3 & 40.8 \\
SAC (ours) & 90.4 & 53.9 & \bfseries 86.6 & \bfseries 42.4 & 27.3 & \bfseries 45.1 & 48.5 & \bfseries 42.7 & \bfseries 87.4 & 40.1 & 86.1 & \bfseries 67.5 & 29.7 & \bfseries 88.5 & \bfseries 49.1 & \bfseries 54.6 & 9.8 & 26.6 & 45.3 & \bfseries 53.8 \\
\bottomrule
\multicolumn{21}{@{}l}{\scriptsize $(^\dagger)$ denotes the use of PSPNet \cite{ZhaoSQWJ17} instead of DeepLabv2 \cite{ChenPKMY18}.} \\
\end{tabularx}
\caption{\textbf{Per-class IoU (\%) comparison} on GTA5 $\rightarrow$ Cityscapes adaptation, evaluated on the Cityscapes validation set.}
\label{table:result_gta_to_city}
\vspace{-0.5em}
\end{table*}

\myparagraph{Setup.}
We adopt the established evaluation protocol from previous work \cite{LianDLG19,TsaiHSS0C18,VuJBCP19}.
The synthetic traffic scenes from GTA5 \cite{RichterVRK16} and SYNTHIA \cite{RosSMVL16} serve as the source data, and the real images from the Cityscapes dataset as the target (obviously ignoring the available semantic labels).
This results in two domain adaptation scenarios depending on the choice of the source data: GTA5 $\rightarrow$ Cityscapes and SYNTHIA $\rightarrow$ Cityscapes.
As in previous work, at training time we only use the training split of the Cityscapes dataset and report the results on the validation split.
We measure the segmentation accuracy with per-class Intersection-over-Union (IoU) and its average, the mean IoU (mIoU).

\subsection{Implementation details}
We implement our framework in PyTorch \cite{NEURIPS2019_9015}.
We adopt DeepLabv2 \cite{ChenPKMY18} as the segmentation architecture, and evaluate our method with two backbones, ResNet-101 \cite{HeZRS16} and VGG16 \cite{SimonyanZ14a}, following recent work \cite{KimB20a,TsaiHSS0C18,TsaiSSC19,VuJBCP19,WangYWFXHHS20}.
Both backbones initialise from the models pre-trained on ImageNet \cite{imagenet_cvpr09}.
We first train the models with ABN \cite{LiWSHL18} (\cf~\cref{sec:training}), implemented via SyncBN \cite{NEURIPS2019_9015}, on multi-scale crops resized to $640 \times 640$ and a batch size of 16.
Next, training proceeds with the self-supervised target loss (\cf~\cref{sec:selftrain}) and the BatchNorm layers \cite{IoffeS15} frozen.
The batch size of $16$ comprises $8$ source images and $8$ target images at resolution $1024 \times 512$, which is a common practice \cite{Wang_2020_ECCV,0001S20}.
The target batch contains only two image samples along with $3$ random crops each (\ie~$N=3$ in \cref{sec:batch}), downscaled up to a factor of $0.5$.
As the photometric noise, we use colour jitter, random blur and greyscaling (see \cref{sec:supp_impl} for details).
The optimisation uses SGD with a constant learning rate of $2.5 \times 10^{-4}$, momentum $0.9$ and weight decay of $5 \times 10^{-4}$.
We accumulate the gradient in alternating source-target forward passes to keep the memory footprint in check.
Since the focal term in \cref{eq:loss} reduces the target loss magnitude \wrt the source loss, we scale it up by a factor of $5$ ($2$ for VGG-16).
We train our VGG-based framework on two TITAN X GPUs (12GB), while the ResNet-based variant requires four.
This is a substantially reduced requirement compared to recent work (\eg, FADA \cite{Wang_2020_ECCV} requires 4 Tesla P40 GPUs with 24GB memory).
Note that the momentum network is always in evaluation mode, has gradient tracking disabled, hence adds only around $35\%$ memory overhead.
For the momentum network, we fix $\gamma_\psi = 0.99$ and $T = 100$ in all our experiments.
For the other hyperparameters, we use $\gamma_\chi = 0.99$, $\zeta = 0.75$, $\beta = 10^{-3}$ and $\lambda = 3$.
\cref{sec:hyper_sensitivity} provides further detail on hyperparameter selection, as well as a sensitivity analysis of our framework \wrt $\zeta$ and $\beta$.
The inference follows the usual procedure of a single forward pass through the segmentation network at the original image resolution without any post-processing.

\subsection{Comparison to state of the art}
We compare our approach to the current state of the art on the two domain adaptation scenarios: GTA5 $\rightarrow$ Cityscapes in \cref{table:result_gta_to_city} and SYNTHIA $\rightarrow$ Cityscapes in \cref{table:synthia_gta_to_city}.
For a fair comparison, all numbers originate from single-scale inference.
In both cases, our approach, denoted as SAC (``Self-supervised Augmentation Consistency''), substantially outperforms our baseline (\ie~the source-only loss model with ABN, see \cref{sec:training}), and, in fact, sets a new state of the art in terms of mIoU.
Importantly, while the ranking of previous works depends on the backbone choice and the source data, we reach the top rank consistently in all settings.

\begin{table*}
\footnotesize
\begin{tabularx}{\linewidth}{@{}X|S[table-format=2.1]@{\hspace{0.8em}}S[table-format=2.1]@{\hspace{0.8em}}S[table-format=2.1]@{\hspace{0.8em}}S[table-format=2.1]@{\hspace{0.8em}}S[table-format=2.1]@{\hspace{0.8em}}S[table-format=2.1]@{\hspace{0.8em}}S[table-format=2.1]@{\hspace{0.8em}}S[table-format=2.1]@{\hspace{0.8em}}S[table-format=2.1]@{\hspace{0.8em}}S[table-format=2.1]@{\hspace{0.8em}}S[table-format=2.1]@{\hspace{0.8em}}S[table-format=2.1]@{\hspace{0.8em}}S[table-format=2.1]@{\hspace{0.8em}}S[table-format=2.1]@{\hspace{0.8em}}S[table-format=2.1]@{\hspace{0.8em}}S[table-format=2.1]@{\hspace{0.8em}}|S[table-format=2.1]@{\hspace{0.8em}}|S[table-format=2.1]@{}}
\toprule
Method & {road} & {sidew} & {build} & {wall} & {fence} & {pole} & {light} & {sign} & {veg} & {sky} & {pers} & {ride} & {car} & {bus} & {moto} & {bicy} & {mIoU$^\text{13}$} & {mIoU} \\
\midrule
\multicolumn{19}{@{}l}{\scriptsize \textit{Backbone: VGG-16}} \\
\midrule
PyCDA \cite{LianDLG19} & 80.6 & 26.6 & 74.5 & 2.0 & 0.1 & 18.1 & 13.7 & 14.2 & 80.8 & 71.0 & 48.0 & 19.0 & 72.3 & 22.5 & 12.1 & 18.1 & \textemdash & 35.9 \\
PIT \cite{LvLCL20} & 81.7 & 26.9 & 78.4 & 6.3 & 0.2 & 19.8 & 13.4 & 17.4 & 76.7 & 74.1 & 47.5 & 22.4 & 76.0 & 21.7 & 19.6 & 27.7 & \textemdash & 38.1 \\
FADA \cite{Wang_2020_ECCV} & 80.4 & 35.9 & 80.9 & 2.5 & 0.3 & 30.4 & 7.9 & 22.3 & \bfseries 81.8 & \bfseries 83.6 & 48.9 & 16.8 & 77.7 & 31.1 & 13.5 & 17.9 & \textemdash & 39.5 \\
FDA \cite{0001S20} & \bfseries 84.2 & 35.1 & 78.0 & 6.1 & 0.4 & 27.0 & 8.5 & 22.1 & 77.2 & 79.6 & 55.5 & 19.9 & 74.8 & 24.9 & 14.3 & 40.7 & \textemdash & 40.5 \\
CD-AM \cite{Yang_2021_WACV} & 73.0 & 31.1 & 77.1 & 0.2 & 0.5 & 27.0 & 11.3 & 27.4 & 81.2 & 81.0 & 59.0 & \bfseries 25.6 & 75.0 & 26.3 & 10.1 & 47.4 & \textemdash & 40.8 \\
LDR \cite{Yang_2020_ECCV} & 73.7 & 29.6 & 77.6 & 1.0 & 0.4 & 26.0 & 14.7 & 26.6 & 80.6 & 81.8 & 57.2 & 24.5 & 76.1 & 27.6 & 13.6 & 46.6 & \textemdash &41.1 \\
SA-I2I \cite{MustoZ20} & 79.1 & 34.0 & 78.3 & 0.3 & 0.6 & 26.7 & 15.9 & \bfseries 29.5 & 81.0 & 81.1 & 55.5 & 21.9 & 77.2 & 23.5 & 11.8 & 47.5 & \textemdash & 41.5 \\
\midrule
Baseline (ours) & 60.7 & 26.9 & 67.1 & 8.3 & 0.0 & 33.5 & 11.9 & 18.3 & 66.4 & 70.4 & 52.1 & 16.1 & 64.6 & 15.5 & 11.5 & 26.4 & 39.1 & 34.4 \\
SAC (ours) & 77.9 & \bfseries 38.6 & \bfseries 83.5 & \bfseries 15.8 & \bfseries 1.5 & \bfseries 38.2 & \bfseries 41.3 & 27.9 & 80.8 & 83.0 & \bfseries 64.3 & 21.2 & \bfseries 78.3 & \bfseries 38.5 & \bfseries 32.6 & \bfseries 62.1 & 56.2 & \bfseries 49.1 \\
\midrule
\multicolumn{19}{@{}l}{\scriptsize \textit{Backbone: ResNet-101}} \\
\midrule
ADVENT \cite{VuJBCP19} & 85.6 & 42.2 & 79.7 & 8.7 & 0.4 & 25.9 & 5.4 & 8.1 & 80.4 & 84.1 & 57.9 & 23.8 & 73.3 & 36.4 & 14.2 & 33.0 & \textemdash & 41.2 \\
PIT \cite{LvLCL20} & 83.1 & 27.6 & 81.5 & 8.9 & 0.3 & 21.8 & 26.4 & \bfseries 33.8 & 76.4 & 78.8 & 64.2 & 27.6 & 79.6 & 31.2 & 31.0 & 31.3 & \textemdash & 44.0 \\
PyCDA$^\dagger$ \cite{LianDLG19} & 75.5 & 30.9 & 83.3 & 20.8 & 0.7 & 32.7 & 27.3 & 33.5 & 84.7 & 85.0 & 64.1 & 25.4 & 85.0 & 45.2 & 21.2 & 32.0 & \textemdash & 46.7 \\
CD-AM \cite{Yang_2021_WACV} & 82.5 & 42.2 & 81.3 & \textemdash & \textemdash & \textemdash & 18.3 & 15.9 & 80.6 & 83.5 & 61.4 & 33.2 & 72.9 & 39.3 & 26.6 & 43.9 & 52.4 & \textemdash \\
FDA \cite{0001S20} & 79.3 & 35.0 & 73.2 & \textemdash & \textemdash & \textemdash & 19.9 & 24.0 & 61.7 & 82.6 & 61.4 & 31.1 & 83.9 & 40.8 & \bfseries 38.4 & 51.1 &52.5 & \textemdash \\
LDR \cite{Yang_2020_ECCV} & 85.1 & 44.5 & 81.0 & \textemdash & \textemdash & \textemdash & 16.4 & 15.2 & 80.1 & 84.8 & 59.4 & \bfseries 31.9 & 73.2 & 41.0 & 32.6 & 44.7 & 53.1 & \textemdash \\
SA-I2I \cite{MustoZ20} & 87.7 & \bfseries 49.7 & 81.6 & \textemdash & \textemdash & \textemdash & 19.3 & 18.5 & 81.1 & 83.7 & 58.7 & 31.8 & 73.3 & \bfseries 47.9 & 37.1 & 45.7 & 55.1 & \textemdash \\
FADA \cite{Wang_2020_ECCV} & 84.5 & 40.1 & 83.1 & 4.8 & 0.0 & 34.3 & 20.1 & 27.2 & 84.8 & 84.0 & 53.5 & 22.6 & 85.4 & 43.7 & 26.8 & 27.8 & \textemdash & 45.2 \\
IAST \cite{Mei_2020_ECCV} & 81.9 & 41.5 & 83.3 & 17.7 & \bfseries 4.6 & 32.3 & 30.9 & 28.8 & 83.4 & 85.0 & \bfseries 65.5 & 30.8 & 86.5 & 38.2 & 33.1 & 52.7 & \textemdash & 49.8 \\
RPT$^\dagger$ \cite{ZhangQYNL020} & 88.9 & 46.5 & 84.5 & 15.1 & 0.5 & 38.5 & 39.5 & 30.1 & 85.9 & 85.8 & 59.8 & 26.1 & \bfseries 88.1 & 46.8 & 27.7 & \bfseries 56.1 & \textemdash & 51.2 \\
\midrule
Baseline (ours) & 63.9 & 25.9 & 71.0 & 11.0 & 0.2 & 36.9 & 7.6 & 20.0 & 72.9 & 75.5 & 46.7 & 16.7 & 74.5 & 15.8 & 20.8 & 21.7 & 41.0 & 36.3 \\
SAC (ours) & \bfseries 89.3 & 47.2 & \bfseries 85.5 & \bfseries 26.5 & 1.3 & \bfseries 43.0 & \bfseries 45.5 & 32.0 & \bfseries 87.1 & \bfseries 89.3 & 63.6 & 25.4 & 86.9 & 35.6 & 30.4 & 53.0 & \bfseries 59.3 & \bfseries 52.6 \\
\bottomrule
\multicolumn{19}{@{}l}{\scriptsize $(^\dagger)$ denotes the use of PSPNet \cite{ZhaoSQWJ17} instead of DeepLabv2 \cite{ChenPKMY18}. \ \ mIoU$^\text{13}$ is the average IoU over 13 classes (\ie~excluding ``wall'', ``fence'' and ``pole'').} \\
\end{tabularx}
\caption{\textbf{Per-class IoU (\%) comparison} on SYNTHIA $\rightarrow$ Cityscapes adaptation, evaluated on the Cityscapes validation set.}
\label{table:synthia_gta_to_city}
\vspace{-0.5em}
\end{table*}

\myparagraph{GTA5 $\rightarrow$ Cityscapes (\cref{table:result_gta_to_city}).}
Our method achieves a clear improvement over the best published results \cite{MustoZ20,ZhangQYNL020} of $+3.4\%$ and $+1.2\%$ using the VGG-16 and ResNet-101 backbones, respectively.
Note that RPT \cite{ZhangQYNL020} and SA-I2I \cite{MustoZ20} have a substantially higher model complexity.
RPT \cite{ZhangQYNL020} uses PSPNet \cite{ZhaoSQWJ17}, which has a higher upper bound than DeepLabv2 in a fully supervised setup (\eg, $+5.7 \%$ IoU on PASCAL VOC \cite{ZhaoSQWJ17}); it requires extracting superpixels and training an encoder-decoder LSTM, thus increasing the model capacity and the computational overhead.
SA-I2I \cite{MustoZ20} initialises from a stronger baseline, BDL \cite{LiYV19}, and relies on a style transfer network and adversarial training.
While both RPT \cite{ZhangQYNL020} and SA-I2I \cite{MustoZ20} require multiple rounds of training, 3 and 6 (from BDL \cite{LiYV19}), respectively, we train with the target loss in a single pass.
Notably, compared to the previous best approach for VGG with a ResNet evaluation, SA-I2I \cite{MustoZ20}, our improvement with ResNet-101 is substantial, $+3.4\%$, and is comparable to the respective margin on VGG-16.

\myparagraph{SYNTHIA $\rightarrow$ Cityscapes (\cref{table:synthia_gta_to_city}).}
Here, the result is consistent with the previous scenario.
Our approach attains state-of-the-art accuracy for both backbones, improving by $7.6\%$ and $1.4\%$ with VGG-16 and ResNet-101 backbones over the best results previously published \cite{MustoZ20,ZhangQYNL020}.
Again, our method with ResNet-101 outperforms the previous best method with full evaluation, PyCDA \cite{LianDLG19}, by $5.9\%$ IoU.

Remarkably, in both settings our approach is more accurate or competitive with many recent works \cite{subhani2020learning,Wang_2020_ECCV,Yang_2020_ECCV} even when using a weaker backbone, \ie~VGG-16 instead of ResNet-101.
This is significant, as these improvements are not due to increased training complexity or model capacity, in contrast to these previous works.
Additional results, including the evaluation on Cityscapes \textit{test}, are shown in \cref{sec:supp_class,sec:supp_eval}.

\begin{table}
\footnotesize
\setlength\fboxsep{0pt}
\begin{tabularx}{\linewidth}{@{}>{\raggedleft\arraybackslash}m{18mm}|S[table-format=2.1]|X@{}}
\toprule
\multicolumn{1}{>{\centering\arraybackslash}m{18mm}|}{$\Delta$} & mIoU & Configuration \\
\midrule
-8.0 \cbar{36}{5} & 41.9 & No augmentation consistency \\
-6.4 \cbar{28.8}{5} & 43.5 & No momentum net ($\gamma_\psi = 0$, $T=1$) \\
-3.9 \cbar{17.6}{5} & 46.0 & No photometric noise \\
-2.6 \cbar{11.7}{5} & 47.3 & No multi-scale fusion \\
-2.4 \cbar{10.8}{5} & 47.5 & No focal loss ($\lambda = 0$) \\
-1.9 \cbar{8.6}{5} & 48.0 & Min. entropy fusion (\vs averaging) \\
-1.7 \cbar{7.7}{5} & 48.2 & No class-based thresholding ($\beta \rightarrow 0$) \\
-1.6 \cbar{7.2}{5} & 48.3 & No confidence regularisation  \\
-1.5 \cbar{6.8}{5} & 48.4 & No importance sampling \\
-0.6 \cbar{2.7}{5} & 49.3 & No horizontal flipping  \\
\toprule
0.0 & 49.9 & Full framework (VGG-16) \\
\bottomrule
\end{tabularx}
\caption{\textbf{Ablation study.} We use the GTA5 $\rightarrow$ Cityscapes setting with the VGG-based model to study the effect of the components of our framework by individually removing each. We report the mean IoU for the Cityscapes validation split.}
\label{table:ablation}
\vspace{-0.5em}
\end{table}

\begin{figure*}[t]
\begin{subfigure}{.02\linewidth}
	\scriptsize
	\vspace{-1.2em}
	\rotatebox{90}{\textbf{Ground truth}}\vspace{3.2em}
	\rotatebox{90}{Adapted $\leftarrow$ \textbf{VGG-16} $\rightarrow$ Baseline}\vspace{2.5em}
	\rotatebox{90}{Adapted $\leftarrow$ \textbf{ResNet-101} $\rightarrow$ Baseline}
\end{subfigure}%
\hspace{0.1em}%
\begin{subfigure}{.49\linewidth}
  \centering
    \includegraphics[width=0.49\linewidth]{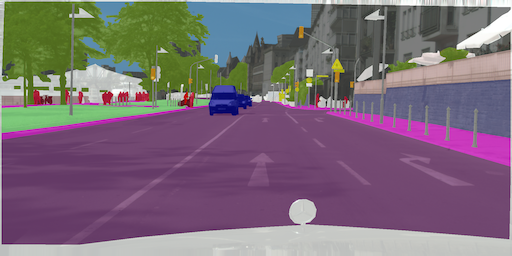}
    \includegraphics[width=0.49\linewidth]{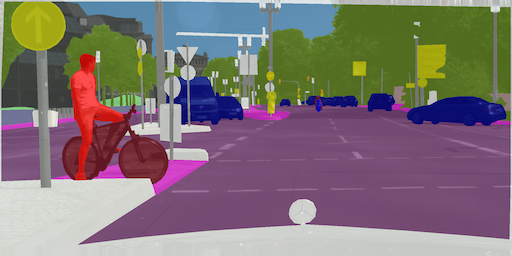}\\
	\includegraphics[width=0.49\linewidth]{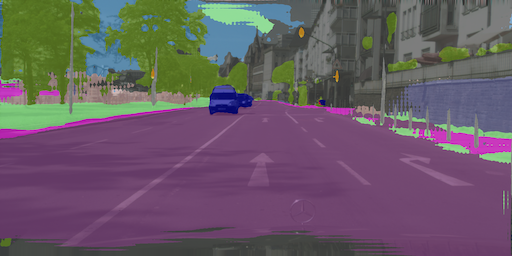}
    \includegraphics[width=0.49\linewidth]{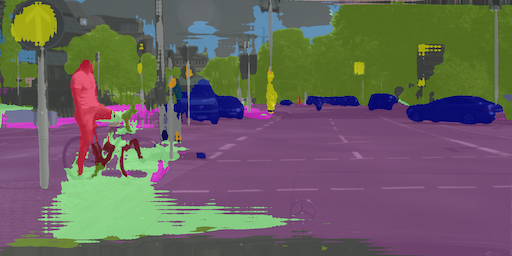}\\
    \includegraphics[width=0.49\linewidth]{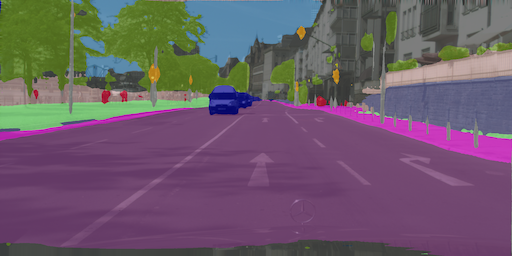}
    \includegraphics[width=0.49\linewidth]{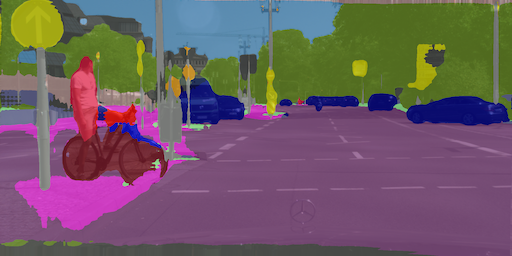}\\
    \includegraphics[width=0.49\linewidth]{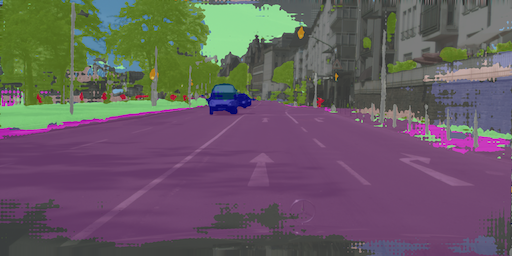}
    \includegraphics[width=0.49\linewidth]{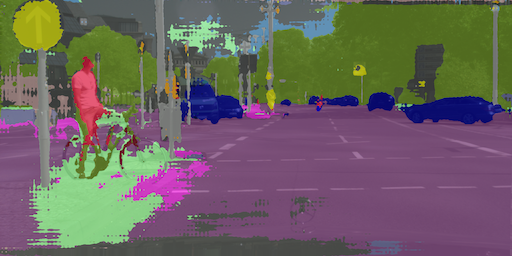}\\
    \includegraphics[width=0.49\linewidth]{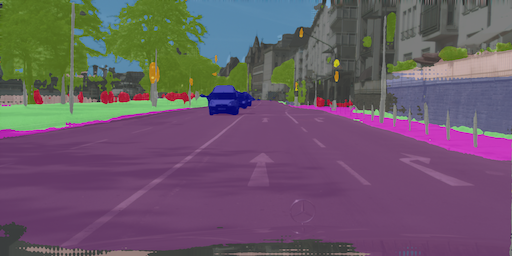}
    \includegraphics[width=0.49\linewidth]{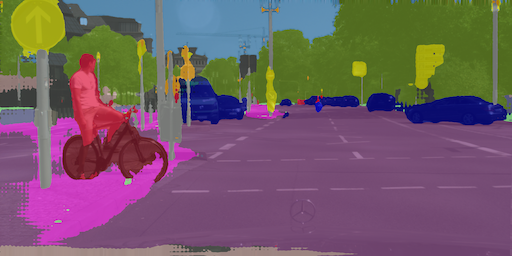}
    \caption{\scriptsize GTA5 $\rightarrow$ Cityscapes}
\end{subfigure}%
\hspace{0.3em}%
\begin{subfigure}{.49\linewidth}
  \centering
    \includegraphics[width=0.49\linewidth]{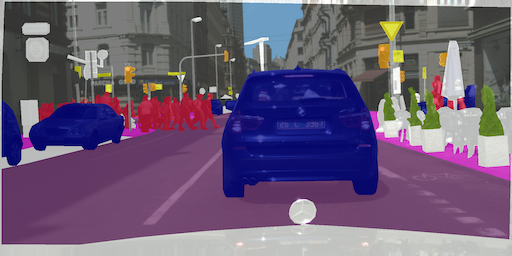}
    \includegraphics[width=0.49\linewidth]{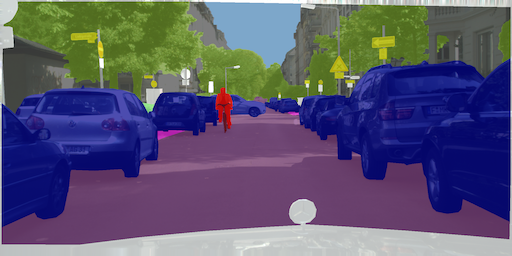}\\
	\includegraphics[width=0.49\linewidth]{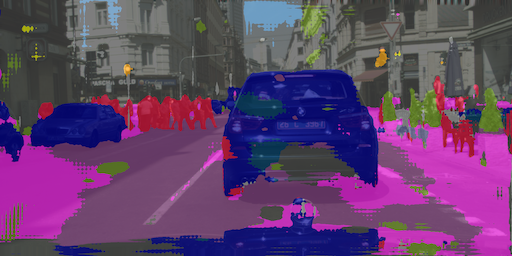}
    \includegraphics[width=0.49\linewidth]{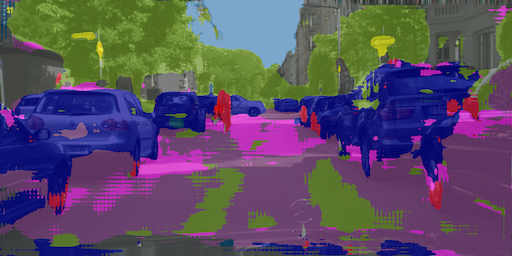}\\
    \includegraphics[width=0.49\linewidth]{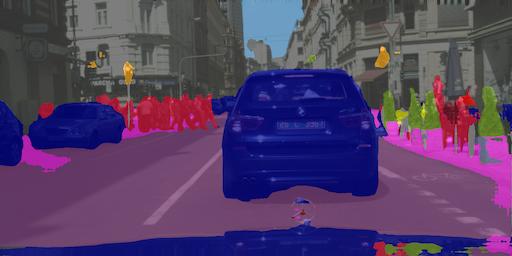}
    \includegraphics[width=0.49\linewidth]{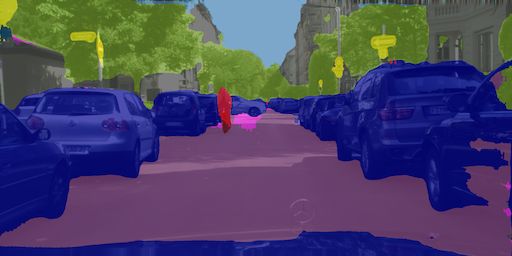}\\
    \includegraphics[width=0.49\linewidth]{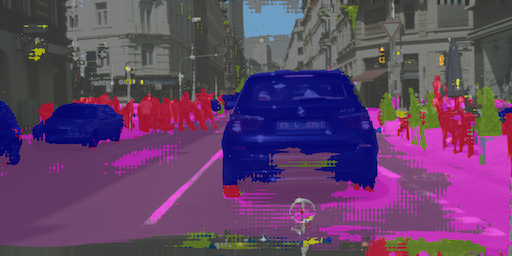}
    \includegraphics[width=0.49\linewidth]{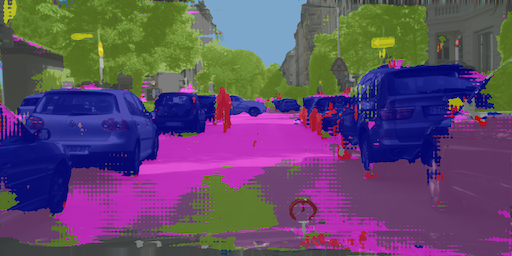}\\
    \includegraphics[width=0.49\linewidth]{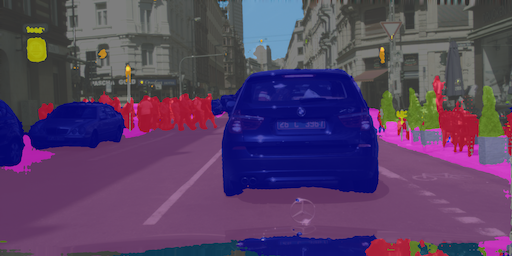}
    \includegraphics[width=0.49\linewidth]{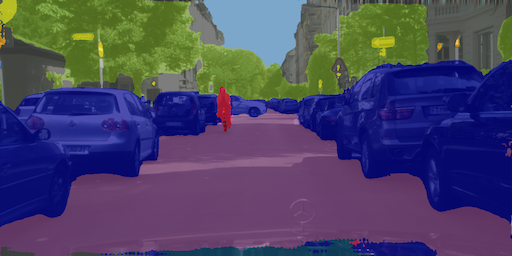}
  \caption{\scriptsize SYNTHIA $\rightarrow$ Cityscapes}
\end{subfigure}
\caption{\textbf{Qualitative examples.} Our approach rectifies an appreciable amount of erroneous predictions from the baseline.}
\label{fig:qualitative}
\vspace{-0.5em}
\end{figure*}

\subsection{Ablation study}
To understand what makes our framework effective, we conduct an ablation study using the GTA5 $\rightarrow$ Cityscapes setting with the VGG-16 backbone.
We independently switch off each component and report the results in \cref{table:ablation}.
We find that two components, augmentation consistency and the momentum network, play a crucial role.
Disabling the momentum network leads to a $6.4 \%$ IoU decrease, while abolishing augmentation consistency leads to a drop of $8.0 \%$ IoU.
Recall that augmentation consistency comprises three augmentation techniques: photometric noise, multi-scale fusion and random flipping.
We further assess their individual contributions.
Training without the photometric jitter deteriorates the IoU more severely, by $3.9 \%$, compared to disabling the multi-scale fusion ($-2.6 \%$) or flipping ($-0.6 \%$).
We hypothesise that encouraging model robustness to photometric noise additionally alleviates the inductive bias inherited from the source domain to rely on strong appearance cues (\eg, colour and texture), which can be substantially different from the target domain.

Following the intuition that high-confidence predictions should be preferred \cite{subhani2020learning}, we study an alternative implementation of the multi-scale fusion.
For overlapping pixels, instead of averaging the predictions, we pool the prediction with the minimum entropy.
The accuracy drop by $1.9\%$ is somewhat expected.
Averaging predictions via data augmentation has previously been shown to produce well-calibrated uncertainty estimates \cite{ayhan2018test}.
This is important for our method, since it relies on the confidence values to select the predictions for use in self-supervision.
Importance sampling contributes $1.5 \%$ IoU to the total accuracy.
This is surprisingly significant despite that our estimates $\chi_{c,l}$ are only approximate (\cf \cref{sec:training}), but the overall benefit is in line with previous work \cite{GuptaDG19}.
Recall from \cref{eq:sbrt} that our confidence thresholds are computed per class to encourage lower values for long-tail classes.
Disabling this scheme is equivalent to setting $\beta \rightarrow 0$ in \cref{eq:sbrt}, which reduces the mean IoU by $1.7 \%$.
This confirms our observation that the model tends to predict lower confidences for the classes occupying only few pixels.
Similarly, the loss in \cref{eq:loss} without the focal term ($\lambda = 0$) and confidence regularisation ($m_{c^\ast\!,n} = 1$) are $2.4 \%$ and $1.6 \%$ IoU inferior.
This is a surprisingly significant contribution at a negligible computational cost.

\subsection{Qualitative assessment}
\cref{fig:qualitative} presents a few qualitative examples, comparing our approach to the naive baseline (\ie~source-only loss with ABN).
Particularly prominent are the refinements of the classes ``road'', ``sidewalk'' and ``sky'', but even small-scale elements improve substantially (\eg, ``person'', ``fence'' in the leftmost column).
This is perhaps not surprising, owing to our multi-scale training and the thresholding technique, which initially ignores incorrectly predicted pixels in self-supervision (as they initially tend to have low confidence).
Remarkably, the segment boundaries tend to align well with the object boundaries in the image, although our framework \emph{has no explicit encoding of spatial priors}, which was previously deemed necessary \cite{Chen0G18,TsaiSSC19,ZhangDG17,ZhangQYNL020}.
We believe that enforcing semantic consistency with data augmentation makes our method less prone to the contextual bias \cite{ShettySF19}, often blamed for coarse boundaries.

\section{Conclusion}

We presented a simple and accurate approach for domain adaptation of semantic segmentation.
With ordinary augmentation techniques and momentum updates, we achieve state-of-the-art accuracy,
yet make no sacrifice of the modest training or model complexity.
No components of our framework are strictly specialised; they build on a relatively weak and broadly applicable assumption (\cf \cref{sec:intro}).
Although this work focuses on semantic segmentation, we are keen to explore the potential of the proposed techniques for adaptation of other dense prediction tasks, such as optical flow, monocular depth, panoptic and instance segmentation, or even compositions of these multiple tasks.

{\small
\myparagraph{Acknowledgements.} This work has been co-funded by the LOEWE initiative (Hesse, Germany) within the emergenCITY center. Calculations for this research were partly conducted on the Lichtenberg high performance computer of TU Darmstadt.
}

{\small
\bibliographystyle{ieee_fullname}
\bibliography{egbib}
}

\clearpage
\pagenumbering{roman}
\appendix

\title{Self-supervised Augmentation Consistency \\for Adapting Semantic Segmentation\\[1mm]\large -- Supplemental Material --}
\author{Nikita Araslanov$^1$ \hspace{1cm} Stefan Roth$^{1,2}$\\
$\ ^1$Department of Computer Science, TU Darmstadt \hspace{1cm} $\ ^2$ hessian.AI}

\maketitle

\section{Overview}
In this appendix, we first provide further training and implementation details of our framework.
We then take a closer look at the accuracy of long-tail classes, before and after adaptation.
Next, we discuss our strategy for hyperparameter selection and perform a sensitivity analysis.
We also evaluate our framework using another segmentation architecture, FCN8s \citesupp{ShelhamerLD17}.
Finally, we discuss the limitations of the current evaluation protocol and propose a revision based on the best practices in the field at large.

\section{Further Technical Details}
\label{sec:supp_impl}

\paragraph{Photometric noise.}
Recall that our framework uses random Gaussian smoothing, greyscaling and colour jittering to implement the photometric noise.
We re-use the parameters for these operations from the MoCo-v2 framework \citesupp{chen2020mocov2}.
In detail, the kernel radius for the Gaussian blur is sampled uniformly from the range $[0.1, 2.0]$.
Note that this does not correspond to the actual filter size.\footnote{The Pillow Library \citesupp{clark2015pillow} internally converts the radius $r$ to the box length as $L = \sqrt{3 * r^2 + 1}$.}
The colour jitter, applied with probability $0.5$, implements a perturbation of the image brightness, contrast and saturation with a factor sampled uniformly from $[0.6, 1.4]$, while the hue factor is sampled uniformly at random in the range of $[0.9, 1.1]$.
We convert a target image to its greyscale version with probability \num{0.2}.
\cref{fig:photometric} demonstrates an example implementation of this procedure in Python.

\begin{figure}[t]
\begin{lstlisting}[language=Python]
import random
import PIL
import torchvision.transforms as tf
import torchvision.transforms.functional as F

# Load the image
image = PIL.Image.open(...)

# Gaussian blur
# with a randomly sampled radius
radius = random.uniform(.1,2.)
gaussian = PIL.ImageFilter.GaussianBlur(radius)
image = image.filter(gaussian)

# Colour jitter
# with probability 0.5
if 0.5 > random.random():
	jitter = tf.ColorJitter(brightness=0.4,
	                        contrast=0.4,
	                        saturation=0.4,
	                        hue=0.1)
	image = jitter(image)

# Convert to greyscale
# with probability 0.2
if 0.2 > random.random():
	image = F.to_grayscale(image)
\end{lstlisting}
\vspace{-0.5em}
\caption{\textbf{Python implementation of the photometric noise.}}
\label{fig:photometric}
\vspace{-0.5em}
\end{figure}

\myparagraph{Constraint-free data augmentation.}
Similarly to the multi-scale cropping of the target images, we scale the source images randomly with a factor sampled uniformly from $[0.5, 1.0]$ prior to cropping.
However, we do not enforce the semantic consistency for the source data, since the ground truth of the source images is available.
For both the target and source images we also use random horizontal flipping.
We additionally experimented with moderate rotation (both with and without semantic consistency), but did not observe a significant effect on the mean accuracy.

\begin{table*}[t!]
\footnotesize
\begin{tabularx}{\linewidth}{@{}>{\centering\arraybackslash}p{1.5em}>{\centering\arraybackslash}p{1.5em}>{\centering\arraybackslash}p{1.5em}|S[table-format=2.1]@{\hspace{0.74em}}S[table-format=2.1]@{\hspace{0.74em}}S[table-format=2.1]@{\hspace{0.74em}}S[table-format=2.1]@{\hspace{0.74em}}S[table-format=2.1]@{\hspace{0.74em}}S[table-format=2.1]@{\hspace{0.74em}}S[table-format=2.1]@{\hspace{0.74em}}S[table-format=2.1]@{\hspace{0.74em}}S[table-format=2.1]@{\hspace{0.74em}}S[table-format=2.1]@{\hspace{0.74em}}S[table-format=2.1]@{\hspace{0.74em}}S[table-format=2.1]@{\hspace{0.74em}}S[table-format=2.1]@{\hspace{0.74em}}S[table-format=2.1]@{\hspace{0.74em}}S[table-format=2.1]@{\hspace{0.74em}}S[table-format=2.1]@{\hspace{0.74em}}S[table-format=2.1]@{\hspace{0.74em}}S[table-format=2.1]@{\hspace{0.74em}}S[table-format=2.1]@{\hspace{0.74em}}|S[table-format=2.1]@{}}
\toprule
CBT & IS & FL & {road} & {sidew} & {build} & {wall} & {fence} & {pole} & {light} & {sign} & {veg} & {terr} & {sky} & {pers} & {ride} & {car} & {truck} & {bus} & {train} & {moto} & {bicy} & {mIoU} \\
\midrule
 & & & 88.1 & 41.0 & 85.7 & 30.8 & 30.6 & 33.1 & 37.0 & 22.9 & 86.6 & 36.8 & 90.7 & 67.1 & 27.1 & 86.8 & 34.4 & 30.4 & 8.5 & 7.5 & 0.0 & 44.5 \\
\midrule
& & \cmark & 89.4 & 52.3 & 86.0 & \bfseries 34.0 & 32.6 & \bfseries 38.5 & 43.3 & 30.6 & 85.2 & 30.9 & 88.5 & 66.7 & 28.0 & 85.7 & 35.6 & 39.6 & 0.0 & 6.6 & 0.0 & 46.0 \\
& \cmark & & \bfseries 90.0 & 47.1 & 85.6 & 31.3 & 24.9 & 32.3 & 38.9 & 28.2 & \bfseries 87.3 & \bfseries 39.8 & 89.4 & \bfseries 67.7 & 28.6 & \bfseries 88.1 & 40.1 & 50.0 & 7.3 & 9.9 & 2.2 & 46.8 \\
\cmark & & & 89.3 & 39.0 & 85.1 & 33.2 & 26.1 & 32.4 & 41.8 & 25.2 & 86.3 & 27.4 & \bfseries 90.4 & 66.4 & 28.2 & 87.5 & 32.9 & 45.4 & 11.0 & 7.6 & 0.0 & 45.0 \\
\midrule
& \cmark & \cmark & 89.3 & 52.6 & 86.0 & 33.4 & 30.0 & 38.0 & 44.9 & 34.3 & 86.9 & 35.3 & 88.0 & 65.4 & 27.3 & 86.2 & 37.6 & 44.0 & 20.9 & 9.6 & 6.5 & 48.2 \\
\cmark & & \cmark & 89.3 & 52.2 & 86.1 & 34.2 & 31.5 & 37.0 & 43.4 & 36.3 & 85.2 & 30.7 & 86.6 & 66.2 & \bfseries 30.3 & 85.3 & 36.2 & 43.9 & \bfseries 29.2 & 6.8 & 8.6 & 48.4 \\
\cmark & \cmark & & 89.7 & 45.1 & 85.6 & 29.6 & 28.3 & 31.7 & 41.9 & 27.5 & 87.2 & 37.4 & 89.8 & 66.9 & 29.2 & 87.5 & 37.3 & 31.6 & 24.7 & 11.9 & 20.2 & 47.5 \\
\midrule
\cmark & \cmark & \cmark & \bfseries 90.0 & \bfseries 53.1 & \bfseries 86.2 & 33.8 & \bfseries 32.7 & 38.2 & \bfseries 46.0 & \bfseries 40.3 & 84.2 & 26.4 & 88.4 & 65.8 & 28.0 & 85.6 & \bfseries 40.6 & \bfseries 52.9 & 17.3 & \bfseries 13.7 & \bfseries 23.8 & \bfseries 49.9 \\
\bottomrule
\end{tabularx}
\caption{\textbf{Per-class IoU (\%)} on Cityscapes \emph{val} using a VGG-16 backbone in the GTA5 $\rightarrow$ Cityscapes setting. We study three components in more detail: class-based thresholding (CBT), importance sampling (IS) and the focal loss (FL). The mIoU of the settings in the last four rows are reproduced from the main text. Here, we elaborate on the per-class accuracy in a broader context of the supplementary experiments in the first four rows.}
\label{table:result_longtail}
\end{table*}

\myparagraph{Training schedule.}
Our framework typically needs $150-200$K iterations in total (\ie~including the source-only pre-training) until convergence, as determined on a random subset of \num{500} images from the training set (see our discussion in \cref{sec:supp_eval} below).
This varies slightly depending on the backbone and the source data used.
This schedule translates to approximately \num{3} days of training with standard GPUs (\eg, Titan X Pascal with 12 GB memory) for both VGG-16 and ResNet-101 backbones.
Recall that we used \num{4} GPUs for our ResNet version of the framework, hence its training time is comparable to the VGG variant, which uses only \num{2} GPUs.
All our experiments use a constant learning rate for simplicity, but more advanced schedules, such as cyclical learning rates \cite{IzmailovPGVW18}, the cosine schedule \citesupp{chen2020mocov2,LoshchilovH17} or ramp-ups \cite{LaineA17}, may further improve the accuracy of our framework.

\section{Additional Experiments}
\label{sec:supp_class}

\subsection{A closer look at long-tail adaptation}
Recall that our framework features three components to attune the adaptation process to the long-tail classes: class-based thresholding (CBT), importance sampling (IS) and the focal loss (FL), which we summarily refer to as the \emph{long-tail components} in the following.
Disabling the long-tail components individually is equivalent to setting $\beta \rightarrow 0$ for CBT, using uniform sampling of the target images instead of IS or assigning $\lambda$ to \num{0} for the FL.
Here, we extend our ablation study of the GTA5 $\rightarrow$ Cityscapes setup with VGG-16 (\cf \cref{table:ablation} from the main text) and experiment with different combinations of the long-tail components.
\cref{table:result_longtail} details the per-class accuracy of the possible compositions.

We observe that the ubiquitous classes -- ``road'', ``building'', ``vegetation'', ``sky'', ``person'' and ``car'' -- are hardly affected;
it is primarily the long-tail categories that change in accuracy.
Furthermore, our long-tail components are mutually complementary.
The mean IoU improves when one of the components is active, from $44.5\%$ to up to $46.8\%$.
It is boosted further with two of the components enabled to $48.4\%$, and reaches its maximum for our model, $49.9\%$, when all three components are in place.

We further identify the following tentative patterns.
FL tends to improve classes ``wall'', ``fence'' and ``pole''.
CBT increases the accuracy of the ``traffic light'' category (which has high image frequency and occupies only a few pixels), but also rare classes, such as ``rider'', ``bus'' and ``train'' benefit from CBT, especially in conjunction with IS.
IS also enhances the mask quality of the classes ``bicycle'' and ``motorcycle''.
Nevertheless, we urge caution against interpreting the results for each class in isolation, despite such widespread practice in the literature.
Today's semantic segmentation models do not possess the notion of an `ambiguous' class prediction and each pixel receives a meaningful label.
By the pigeon's hole principle, this implies that the changes in the IoU of one class have an immediate effect on the IoU of the other classes.
Therefore, the benefits of individual framework components have to be understood in the context of their aggregated effect on multiple classes, \eg~using the mean IoU.
For instance, consider the class ``train'' for which IS appears to also decrease the IoU: while CBT together with FL achieve $29.2\%$ IoU, adding IS decreases the IoU to $17.3\%$.
However, the IoU of other classes increases (\eg, ``motorcycle'', ``bicycle''), as does the mean IoU.
Furthermore, only few classes reach their maximum accuracy when we enable all three long-tail components.
Yet, it is the setting with the best \emph{accuracy trade-off} between the individual classes, \ie~with the highest mean IoU.
Overall, the long-tail components improve our framework by $5.4\%$ mean IoU combined, a substantial margin.

\begin{table*}[t!]
\footnotesize
\begin{tabularx}{\linewidth}{@{}l|S[table-format=2.1]@{\hspace{0.7em}}S[table-format=2.1]@{\hspace{0.7em}}S[table-format=2.1]@{\hspace{0.7em}}S[table-format=2.1]@{\hspace{0.7em}}S[table-format=2.1]@{\hspace{0.7em}}S[table-format=2.1]@{\hspace{0.7em}}S[table-format=2.1]@{\hspace{0.7em}}S[table-format=2.1]@{\hspace{0.7em}}S[table-format=2.1]@{\hspace{0.7em}}S[table-format=2.1]@{\hspace{0.7em}}S[table-format=2.1]@{\hspace{0.7em}}S[table-format=2.1]@{\hspace{0.7em}}S[table-format=2.1]@{\hspace{0.7em}}S[table-format=2.1]@{\hspace{0.7em}}S[table-format=2.1]@{\hspace{0.7em}}S[table-format=2.1]@{\hspace{0.7em}}S[table-format=2.1]@{\hspace{0.7em}}S[table-format=2.1]@{\hspace{0.7em}}S[table-format=2.1]@{\hspace{0.7em}}|c@{}}
\toprule
Method & {road} & {sidew} & {build} & {wall} & {fence} & {pole} & {light} & {sign} & {veg} & {terr} & {sky} & {pers} & {ride} & {car} & {truck} & {bus} & {train} & {moto} & {bicy} & {mIoU} \\
\midrule
\multicolumn{21}{@{}l}{\scriptsize \textit{GTA5 $\rightarrow$ Cityscapes}} \\
\midrule
Baseline (ours) & 76.7 & 28.2 & 74.4 & 12.7 & 19.0 & 27.2 & 28.7 & 12.2 & 77.0 & 18.0 & 70.6 & 54.8 & 20.6 & 79.6 & 19.0 & 19.2 & 20.6 & 27.9 & 11.2 & 36.7 { \scriptsize{(37.1)}} \\
SAC-FCN (ours) & 86.3 & 45.6 & 84.4 & 30.3 & 27.1 & 24.8 & 42.8 & 35.2 & 86.9 & 39.7 & 88.0 & 62.3 & 32.1 & 84.1 & 28.4 & 43.7 & 31.9 & 29.4 & 45.8 & 49.9 { \scriptsize{(49.9)}} \\
\midrule
\multicolumn{21}{@{}l}{\scriptsize \textit{SYNTHIA $\rightarrow$ Cityscapes}} \\
\midrule
Baseline (ours) & 50.7 & 23.8 & 60.9 & 1.8 & 0.1 & 27.7 & 10.5 & 15.7 & 60.1 & \textemdash & 72.4 & 50.1 & 16.0 & 66.5 & \textemdash & 13.7 & \textemdash & 8.5 & 26.8 & 31.6 { \scriptsize{(34.4)}} \\
SAC-FCN (ours) & 74.7 & 34.2 & 81.4 & 19.8 & 1.9 & 27.2 & 34.8 & 27.2 & 80.0 & \textemdash & 86.3 & 61.5 & 20.8 & 82.5 & \textemdash & 31.2 & \textemdash & 32.0 & 53.9 & 46.8 { \scriptsize{(49.1)}} \\
\bottomrule
\end{tabularx}
\caption{\textbf{Per-class IoU (\%)} on Cityscapes \emph{val} using VGG-16 with FCN8s. For reference, the numbers in parentheses in the last column report the mean IoU of the DeepLabv2 architecture (\cf \cref{table:result_gta_to_city,table:synthia_gta_to_city} from the main text).}
\label{table:result_fcn}
\end{table*}

\begin{table}[t]
\centering
\setlength{\tabcolsep}{0.8em}%
\begin{tabularx}{0.7\linewidth}{@{}X@{}S[table-format=2.1]S[table-format=2.1]S[table-format=2.1]@{}}
\toprule
$\downarrow \zeta \hfill/\hfill \beta \rightarrow\;$ & {$0.0001$} & {$0.001$} & {$0.01$} \\
\midrule
$0.7$ & 47.9 & 48.8 & 46.7 \\
$0.75$ & 48.6 & 49.9 & 46.3 \\
$0.8$ & 48.2 & 49.8 & 45.6 \\
\bottomrule
\end{tabularx}
\caption{\textbf{Mean IoU (\%) on GTA5 $\rightarrow$ Cityscapes (val) with varying $\zeta$ and $\beta$.} Our framework maintains strong accuracy under different settings of $\zeta$ and $\beta$. Even with a poor choice (\eg, $\zeta = 0.8$, $\beta = 0.01$), it fares well \wrt the state of the art and outperforms many previous works (\cf \cref{table:result_gta_to_city} from the main text).}
\label{table:sensitivity}
\end{table}

\subsection{Hyperparameter search and sensitivity}
\label{sec:hyper_sensitivity}

To select $\zeta$ and $\beta$, we first experimented with a few reasonable choices ($\zeta \in (0.7, 0.8)$, $\beta \in (0.0001, 0.01)$)\footnote{While $\zeta$ may seem more interpretable (the maximum confidence threshold), a reasonable range for $\beta$ can be derived from $\chi_c$ for the long-tail classes, which is simply the fraction of pixels these classes tend to occupy in the image (see Eq.~3).} using a more lightweight backbone (MobileNetV2 \citesupp{Sandler2018:MIR}).
To measure performance, we use the mean IoU on the validation set (500 images from Cityscapes \textit{train}, as in the main text).

Here, we study our framework's sensitivity to the particular choice of $\zeta$ and $\beta$.
To make the results comparable to our previous experiments, we use VGG-16 and report the mean IoU on Cityscapes \textit{val} in \cref{table:sensitivity}.
We observe moderate deviation of the IoU \wrt $\zeta$.
A more tangible drop in accuracy with $\beta = 0.01$ is expected, as it leads to low-confidence predictions, which are likely to be inaccurate, to be included into the pseudo label.
We note that while a suboptimal choice of these hyperparameters leads to inferior results (with a standard deviation of $\pm 1.4$\% mIoU), even the weakest model with $\zeta = 0.8$ and $\beta = 0.01$ did not fail to considerably improve over the baseline (by $8.5$\% IoU, \cf \cref{table:result_gta_to_city} in the main text).

\begin{table*}[t!]
\footnotesize
\begin{tabularx}{\linewidth}{@{}l|S[table-format=2.1]@{\hspace{0.6em}}S[table-format=2.1]@{\hspace{0.6em}}S[table-format=2.1]@{\hspace{0.6em}}S[table-format=2.1]@{\hspace{0.6em}}S[table-format=2.1]@{\hspace{0.6em}}S[table-format=2.1]@{\hspace{0.6em}}S[table-format=2.1]@{\hspace{0.6em}}S[table-format=2.1]@{\hspace{0.6em}}S[table-format=2.1]@{\hspace{0.6em}}S[table-format=2.1]@{\hspace{0.6em}}S[table-format=2.1]@{\hspace{0.6em}}S[table-format=2.1]@{\hspace{0.6em}}S[table-format=2.1]@{\hspace{0.6em}}S[table-format=2.1]@{\hspace{0.6em}}S[table-format=2.1]@{\hspace{0.6em}}S[table-format=2.1]@{\hspace{0.6em}}S[table-format=2.1]@{\hspace{0.6em}}S[table-format=2.1]@{\hspace{0.6em}}S[table-format=2.1]@{\hspace{0.6em}}|c@{}}
\toprule
Method & {road} & {sidew} & {build} & {wall} & {fence} & {pole} & {light} & {sign} & {veg} & {terr} & {sky} & {pers} & {ride} & {car} & {truck} & {bus} & {train} & {moto} & {bicy} & {mIoU} \\
\midrule
\multicolumn{21}{@{}l}{\scriptsize \textit{GTA5 $\rightarrow$ Cityscapes}} \\
\midrule
SAC-FCN (ours) & 87.5 & 45.2 & 85.0 & 29.2 & 26.4 & 23.3 & 44.2 & 32.0 & 88.3 & 52.6 & 91.2 & 65.2 & 35.0 & 86.0 & 24.4 & 32.8 & 31.4 & 36.9 & 40.5 & 50.4 { \scriptsize{(49.9)}} \\
SAC-VGG (ours) & 91.5 & 53.9 & 86.6 & 34.1 & 31.5 & 36.8 & 47.2 & 36.9 & 85.1 & 38.0 & 91.1 & 68.7 & 31.9 & 87.4 & 31.0 & 46.7 & 22.6 & 24.2 & 24.0 & 51.0 { \scriptsize{(49.9)}} \\
SAC-ResNet (ours) & 91.8 & 54.3 & 87.4 & 36.2 & 30.2 & 43.7 & 49.7 & 42.1 & 89.3 & 54.3 & 90.5 & 71.8 & 34.9 & 89.8 & 38.8 & 47.3 & 24.9 & 38.3 & 43.8 & 55.7 { \scriptsize{(53.8)}} \\
\midrule
\multicolumn{21}{@{}l}{\scriptsize \textit{SYNTHIA $\rightarrow$ Cityscapes}} \\
\midrule
SAC-FCN (ours) & 66.9 & 25.9 & 80.8 & 12.1 & 2.0 & 24.4 & 37.1 & 27.5 & 78.8 & \textemdash & 88.9 & 63.9 & 25.0 & 84.7 & \textemdash & 27.4 & \textemdash & 36.9 & 50.2 & 45.8 { \scriptsize{(46.8)}} \\
SAC-VGG (ours) & 70.4 & 29.7 & 83.6 & 11.6 & 1.8 & 34.2 & 41.2 & 29.2 & 81.0 & \textemdash & 87.1 & 67.9 & 25.4 & 75.9 & \textemdash & 34.3 & \textemdash & 42.5 & 57.5 & 48.3 { \scriptsize{(49.1)}} \\
SAC-ResNet (ours) & 87.4 & 41.0 & 85.5 & 17.5 & 2.6 & 40.5 & 44.7 & 34.4 & 87.9 & \textemdash & 91.2 & 68.0 & 31.0 & 89.3 & \textemdash & 33.2 & \textemdash & 38.6 & 49.9 & 52.7 { \scriptsize{(52.6)}} \\
\midrule
\multicolumn{21}{@{}l}{\scriptsize \textit{Fully supervised (Cityscapes)}} \\
\midrule
DeepLab-ResNet \cite{ChenPKMY18} & 97.9 & 81.3 & 90.4 & 48.8 & 47.4 & 49.6 & 57.9 & 67.3 & 91.9 & 69.4 & 94.2 & 79.8 & 59.8 & 93.7 & 56.5 & 67.5 & 57.5 & 57.7 & 68.8 & 70.4 \\
FCN-VGG \citesupp{ShelhamerLD17} & 97.4 & 78.4 & 89.2 & 34.9 & 44.2 & 47.4 & 60.1 & 65.0 & 91.4 & 69.3 & 93.9 & 77.1 & 51.4 & 92.6 & 35.3 & 48.6 & 46.5 & 51.6 & 66.8 & 65.3 \\
\bottomrule
\end{tabularx}
\caption[\textbf{Per-class IoU (\%)} on Cityscapes \emph{test}]{\textbf{Per-class IoU (\%)} on Cityscapes \emph{test}. In the last column, the numbers in parentheses report the mean IoU on Cityscapes \emph{val} from the previous evaluation scheme (\cf \cref{table:result_gta_to_city,table:synthia_gta_to_city} from the main text) for reference. SAC-FCN denotes our VGG-based model with FCN8s \citesupp{ShelhamerLD17} from \cref{sec:fcn}.}
\label{table:result_city_test}
\end{table*}

\subsection{VGG-16 with FCN8s}
\label{sec:fcn}
A number of previous works (\eg, \cite{MustoZ20,Yang_2020_ECCV,0001S20}) used the FCN8s \citesupp{ShelhamerLD17} architecture with VGG-16, as opposed to DeepLabv2 \cite{ChenPKMY18}, adopted in other works (\eg, \cite{KimB20a,Wang_2020_ECCV}) and ours.
Such architecture exchange appears to have been dismissed in previous work as minor, which used only one of the architectures in the experiments.
However, the segmentation architecture alone may contribute to the observed differences in accuracy of the methods and, more critically, to the improvements otherwise attributed to other aspects of the approach.
To facilitate such transparency in our work, we replace DeepLabv2 with its FCN8s counterpart in our framework (with the VGG-16 backbone) and repeat the adaptation experiments from \cref{sec:exp}, \ie~using two source domains, GTA5 and SYNTHIA, and Cityscapes as the target domain.
We keep the values of the hyperparameters the same, with an exception of the learning rate, which we increase by a factor of \num{2} to $5\times 10^4$.
\cref{table:result_fcn} reports the results of the adaptation, which clearly show that our framework generalises well to other segmentation architectures.
Despite the FCN8s baseline model (source-only loss with ABN) achieving a slightly inferior accuracy compared to DeepLabv2 (\eg, $31.6\%$ \vs~$34.4\%$ IoU for SYNTHIA $\rightarrow$ Cityscapes), our self-supervised training still attains a remarkably high accuracy, $46.8\%$ IoU (\vs~$49.1\%$ with DeepLabv2).
This is substantially higher than the previous best method using FCN8s with the VGG-16 backbone, SA-I2I \cite{MustoZ20}: $+3.4\%$ on GTA5 $\rightarrow$ Cityscapes and $+5.3\%$ on SYNTHIA $\rightarrow$ Cityscapes.

\section{Towards Best-practice Evaluation}
\label{sec:supp_eval}

The current strategy to evaluate domain adaptation (DA) methods for semantic segmentation is to use the ground truth of \num{500} randomly selected images from the Cityscapes \textit{train} split for model selection and to report the final model accuracy on the \num{500} Cityscapes \textit{val} images \cite{LianDLG19}.
In this work, we adhered to this procedure to enable a fair comparison to previous work.
However, this evaluation approach is evidently in discord with the established best practice in machine learning and with the benchmarking practice on Cityscapes \cite{CordtsORREBFRS16}, in particular.

The test set is holdout data to be used only for an unbiased performance assessment (\eg, segmentation accuracy) of the final model \citesupp{0082591}.
While it is conceivable to consult the test set for verifying a number of model variants, such access cannot be unrestrained.
This is infeasible to ensure when the test set annotation is public, as is the case with Cityscapes \textit{val}, however.
Benchmark websites traditionally enable a restricted access to the test annotation through impartial submission policies (\eg, limited number of submissions per time window and user), and Cityscapes officially provides one.\footnote{\url{https://www.cityscapes-dataset.com}}

We, therefore, suggest a simple revision of the evaluation protocol for evaluating future DA methods.
As before, we use Cityscapes \textit{train} as the training data for the target domain, naturally without the ground truth.
For model selection, however, we use Cityscapes \textit{val} images with the ground-truth labels.
The holdout test set for reporting the final segmentation accuracy after adaptation becomes Cityscapes \textit{test}, with the results obtained via submitting the predicted segmentation masks to the official Cityscapes benchmark server.

An additional advantage of this strategy is a clear interpretation of the final accuracy in the context of fully supervised methods that routinely use the same evaluation setup.
Also note that Cityscapes \textit{val} contains images from different cities than Cityscapes \textit{train} (which are also different from Cityscapes \textit{test}).
Therefore, it is more suitable for detecting cases of model overfitting on particularities of the city, since the validation set was previously a subset of the training images.

For future reference, we evaluate our framework (both the DeepLabv2 and FCN8s variants) in the proposed setup and report the results in \cref{table:result_city_test}.
To ease the comparison, we juxtapose our validation results reported in the main text (from \cref{table:result_fcn} for FCN8s).\footnote{To our best knowledge, no previous work published their results in this evaluation setting before.}
As we did not finetune our method to Cityscapes \emph{val} following the previous evaluation protocol, we expect the test accuracy on Cityscapes \emph{test} to be on a par with our previously reported accuracy on Cityscapes \emph{val}.
The results in \cref{table:result_city_test} clearly confirm this expectation: the segmentation accuracy on Cityscapes \emph{test} is comparable to the accuracy on Cityscapes \emph{val} (SYNTHIA $\rightarrow$ Cityscapes) or even tangibly higher (GTA5 $\rightarrow$ Cityscapes).
We remark that the remaining accuracy gap to the fully supervised model is still considerable (70.4\% \vs 55.7\% IoU achieved by our best DeepLabv2 model and 65.3\% \vs 51.0\% IoU compared to our best FCN8s variant), which invites further effort from the research community.

We hope that future UDA methods for semantic segmentation will follow suit in reporting the results on Cityscapes \emph{test}.
Owing to the regulated access to the test set, we believe this setting to offer more transparency and fairness to the benchmarking process, and will successfully drive the progress of UDA for semantic segmentation, as it has done in the past for the fully supervised methods.

{\small
\bibliographystylesupp{ieee_fullname}
\bibliographysupp{supp_egbib}
}

\end{document}